\documentclass[5p,times,authoryear]{elsarticle}
\usepackage{xcolor}
\usepackage{makecell}
\usepackage{subfigure}
\usepackage{enumitem}
\usepackage{url}
\usepackage{graphicx}
\usepackage{multirow}
\usepackage{cite}
\usepackage{amssymb}
\usepackage{caption}
\usepackage[colorlinks=true,linkcolor=green]{hyperref}
\usepackage{amssymb}
\journal{arXiv}%{Knowledge-Based Systems}

\begin{document}
\begin{sloppypar}
\begin{frontmatter}

%% Title, authors and addresses

%% use the tnoteref command within \title for footnotes;
%% use the tnotetext command for theassociated footnote;
%% use the fnref command within \author or \address for footnotes;
%% use the fntext command for theassociated footnote;
%% use the corref command within \author for corresponding author footnotes;
%% use the cortext command for theassociated footnote;
%% use the ead command for the email address,
%% and the form \ead[url] for the home page:
%% \title{Title\tnoteref{label1}}
%% \tnotetext[label1]{}
%% \author{Name\corref{cor1}\fnref{label2}}
%% \ead{email address}
%% \ead[url]{home page}
%% \fntext[label2]{}
%% \cortext[cor1]{}
%% \address{Address\fnref{label3}}
%% \fntext[label3]{}

\title{Unsupervised Spatial-Temporal Feature Enrichment
	and Fidelity Preservation Network for Skeleton
	based Action Recognition}

%% use optional labels to link authors explicitly to addresses:
%% \author[label1,label2]{}
%% \address[label1]{}
%% \address[label2]{}

\author[mymainaddress1]{Chuankun Li}
\ead{chuankun@nuc.edu.cn}
\address[mymainaddress1]{School of Information and Communication Engineering, North University of China}

\author[mymainaddress]{Shuai Li\corref{mycorrespondingauthor}}
\cortext[mycorrespondingauthor]{Corresponding author}
\ead{shuaili@sdu.edu.cn}

\author[mymainaddress]{Yanbo Gao}
\ead{ybgao@sdu.edu.cn}
\address[mymainaddress]{School of Control Science and Engineering and School of Software,  Shandong University, Jinan 250100, China.}

\author[mymainaddress1]{Ping Chen}
\ead{chenping@nuc.edu.cn}

\author[mymainaddress1]{Jian Li}
\ead{lijian@nuc.edu.cn}

\author[mysecondary]{Wanqing Li}
\ead{wanqing@uow.edu.au}

\address[mysecondary]{Advanced Multimedia Research Lab, University of Wollongong, Australia}

%\author[mysecondaryaddress2]{Danchi Jiang}
%\ead{danchi.jiang@utas.edu.au}
%\address[mysecondaryaddress2]{School of Engineering, University of Tasmania}
\begin{abstract}
Unsupervised skeleton based action recognition has achieved remarkable progress recently. Existing unsupervised learning methods suffer from severe overfitting problem, and thus small networks are used, significantly reducing the representation capability. To address this problem, the overfitting mechanism behind the unsupervised learning for skeleton based action recognition is first investigated. It is observed that the skeleton is already a relatively high-level and low-dimension feature, but not in the same manifold as the features for action recognition. Simply applying the existing unsupervised learning method may tend to produce features that discriminate the different samples instead of action classes, resulting in the overfitting problem. To solve this problem, this paper presents an Unsupervised spatial-temporal Feature Enrichment and Fidelity Preservation framework (U-FEFP) to generate rich distributed features that contain all the information of the skeleton sequence. A spatial-temporal feature transformation subnetwork is developed using spatial-temporal graph convolutional network and graph convolutional gate recurrent unit network as the basic feature extraction network. The unsupervised Bootstrap Your Own Latent based learning is used to generate rich distributed features and the unsupervised pretext task based learning is used to preserve the information of the skeleton sequence. The two unsupervised learning ways are collaborated as U-FEFP to produce robust and discriminative representations. Experimental results on three widely used benchmarks, namely NTU-RGB+D-60, NTU-RGB+D-120 and PKU-MMD dataset, demonstrate that the proposed U-FEFP achieves the best performance compared with the state-of-the-art unsupervised learning methods. t-SNE illustrations further validate that U-FEFP can learn more discriminative features for unsupervised skeleton based action recognition.
\end{abstract}

%%Graphical abstract
%\begin{graphicalabstract}
%\includegraphics{grabs}
%\end{graphicalabstract}

%%Research highlights
%\begin{highlights}
%\item Research highlight 1
%\item Research highlight 2
%\end{highlights}


\end{frontmatter}

%% \linenumbers

%% main text
\section{Introduction}
Action recognition using different modalities~\citep{9795869,zyer2021HumanAR} (e.g., video, skeleton)~\citep{Li2023ExploringID,Song2021LearningTR,Yang2021HierarchicalSQ,Hu2020JointLI,Zhang2020DeepMT,WANG201843} has been widely studied due to its wide use in many potential applications such as  autonomous driving and video surveillance. Compared with the conventional RGB video, 3D skeleton owning high-level representation is light-weight and robust to both view differences and complicated background. Therefore, 3D skeleton based action recognition has been widely investigated with methods based on handcrafted features~\citep{Weng2017SpatioTemporalNN,Xia2012ViewIH}, Convolutional Neural Networks (CNNs)~\citep{Ke2017SkeletonNetMD,Ke2017ANR,Li2017JointDM,Hou2018SkeletonOS,Li2019MultiviewBased3A,Xu2018EnsembleOC,Li2022FrequencydrivenCA}, Recurrent Neural Networks (RNNs)~\citep{Li2018IndependentlyRN,Li2019DeepIR, Liu2018SkeletonBasedAR,Song2017AnES} and Graph Convolutional Networks (GCNs)~\citep{yan2018spatial,Shi2019TwoStreamAG,Ye2020DynamicGC,Zhang2020SemanticsGuidedNN,Shi2019SkeletonBasedAR,Kong2022MTTMT,Gao2021SkeletonBasedAR, Liu2022SpatialFA,Peng2021SpatialTG}. However, these methods are developed in a fully supervised manner and require extensive annotated labels, which is expensive and time-consuming. Learning general features from unlabelled data for 3D skeleton based action recognition is still an open problem and highly desired. 

There are two main approaches for unsupervised skeleton based action recognition. One is to utilize an encoder-decoder network and generate useful features by pretext tasks such as auto-regression~\citep{Su2020PREDICTC}, reconstruction~\citep{Zheng2018UnsupervisedRL} and jigsaw puzzle~\citep{Lin2020MS2LMS}. This approach exploits low-level feature representation, and the performance of the downstream task is dependent on the design of pretext tasks. Existing methods~\citep{Su2020PREDICTC,Zheng2018UnsupervisedRL,Lin2020MS2LMS} usually take advantage of the RNNs to encode the input skeleton sequence and then regressively predict them.
%In order to solve this problem, the contrastive learning loss~\citep{Lin2020MS2LMS} is used to optimize the encoder network. However, current unsupervised skeleton based methods~\citep{Su2020PREDICTC,Zheng2018UnsupervisedRL,Lin2020MS2LMS} use the RNNs to capture the temporal information and high-level semantic relationship among joints is ignored. Consequently, they cannot extract effective spatio-temporal representation and usually cannot achieve good performance in unsupervised action recognition.
The second approach is to utilize the contrastive learning such as Bootstrap Your Own Latent (BYOL)~\citep{Grill2020BootstrapYO}, Momentum contrast~\citep{He2020MomentumCF} and exploit the discriminative features among samples in latent space. The methods~\citep{Rao2021AugmentedSB,Li20213DHA,Thoker2021SkeletonContrastive3A, Wang2022ContrastReconstructionRL} learn features by pulling or pushing the features of different samples as positive and negative pairs. %However, these methods usually require a memory bank or large batches which increases GPU memory and computation. Furthermore, positive samples may be regarded as negative pairs in one memory bank which affects the recognition performance. In order to solve these problems, methods~\citep{Zhang2022UnsupervisedSA,Zhang2022SkeletalTU} adopt deep framework including online network and target network and design different objective functions to improve performance. The performance is severely affected by overfitting, even only using fewer convolution kernels in the online network. 
Putting aside their individual problems such as designing relevant task and differentiating positive and negative pairs, both approaches suffer from severe overfitting. The existing networks used in supervised learning ~\citep{Shi2019TwoStreamAG,Ye2020DynamicGC,Zhang2020SemanticsGuidedNN,Shi2019SkeletonBasedAR,Kong2022MTTMT, Gao2021SkeletonBasedAR, Liu2022SpatialFA, Peng2021SpatialTG} cannot work effectively in the unsupervised learning due to this severe overfitting. Consequently, the existing unsupervised learning methods employ very simple models, either using only basic RNN models or using very small models  with fewer neurons~\citep{Zhang2022UnsupervisedSA,Zhang2022SkeletalTU}. However, such simple models with low-dimension features are not capable for the high-level action recognition task, and thus cannot achieve high performance. 

Currently, there is no investigation on the mechanism behind the severe overfitting problem in the unsupervised learning for skeleton based action recognition. In this paper, we first study the overfitting mechanism in the unsupervised skeleton based action recognition learning and show that the existing unsupervised learning method cannot effectively generate features that are highly relevant and useful for action recognition. With skeleton sequences already being relatively high-level and low-dimension representations, the encoder-decoder  architecture and the contrastive learning can easily generate features representing or differentiating each skeleton sequence, but the features may not be useful for the action recognition task. This can be intuitively understood since the high-level skeleton is not in the same manifold as the high-level features for action recognition (considering the example that directly using one fully connected layer cannot produce high action recognition performance). This is further illustrated in the following Motivation Section.

To address the above problem, we propose an Unsupervised spatial-temporal Feature Enrichment and Fidelity Preservation framework (U-FEFP). The proposed network generates \textit{rich distributed spatial-temporal} features \textit{containing all information of the original skeleton}. Our contributions can be summarized as follows. %not only rich distributed high-level spatial-temporal features, but also ensures the features to preserve the fidelity of the original skeleton sequences

\begin{itemize}
	\item  We investigate the mechanism behind the severe overfitting problem in the unsupervised learning for skeleton based action recognition. It is found that features representing each skeleton may not be aligned with the features for action recognition, leading to the requirement of learning rich distributed features. To the best of our knowledge, this is the first research that investigates the overfitting mechanism in the unsupervised skeleton based action recognition.

	%The mechanism behind the severe overfitting problem in the unsupervised learning for skeleton based action recognition is investigated. We illustrate that features representing or discriminating each skeleton may not be aligned with the features for action recognition. Accordingly, we propose to generate \textit{rich distributed spatial-temporal} features \textit{containing all information of the original skeleton} in unsupervised learning. To the best of our knowledge, this is the first research that investigates the overfitting mechanism in the  unsupervised skeleton based action recognition.
	
	\item  Based on our observation on the overfitting mechanism, we develop an unsupervised spatial-temporal feature enrichment and fidelity preservation framework (U-FEFP), which can learn rich distributed features while preserving the fidelity of the original skeleton sequence.
	% Based on our observation on the overfitting mechanism, an unsupervised spatial-temporal feature enrichment and fidelity preservation network (UST-FEFP) is developed by taking advantage of the BYOL based learning and pretext task based learning. UST-FEFP can learn rich distributed features while preserving the fidelity of the original skeleton sequence. 
	
	\item A spatial-temporal feature transformation subnetwork is developed by combining the spatial-temporal graph convolution network (ST-GCN) and the graph convolutional GRU network (GConv-GRU). It can effectively learn the spatial-temporal features with a relatively small model, which further reduces the overfitting problem.
	
	\item  Exhaustive experiments on NTU-RGB+D-60~\citep{shahroudy2016ntu}, NTU-RGB+D-120 ~\citep{Liu2020NTUR1} and PKU-MMD~\citep{Liu2020ABD} datasets verify the capacity of the representations learned by our U-FEFP. It achieves state-of-the-art results under both the unsupervised and semi-supervised training.
	
\end{itemize}

The rest of this paper is organized as follows. Section \ref{sec_relatedwork} briefly describes the related work in the skeleton based action recognition, including representative supervised and unsupervised methods. Section \ref{sec_motivation} illustrates the motivation of this paper, and the proposed method is shown in Section \ref{sec_proposed}. Experimental results are presented in Section \ref{sec_exp} with detailed ablation study, and Section \ref{sec_conclusion} draws the conclusion. 

%This letter is structured as follows: The proposed Ishift-GCN is introduced in Section~\uppercase\expandafter{\romannumeral2}. The experiment  evaluation is shown in Section~\uppercase\expandafter{\romannumeral3}. The conclusion and future work are presented in Section~\uppercase\expandafter{\romannumeral4}.
% needed in second column of first page if using \IEEEpubid
%\IEEEpubidadjcol

\vspace{3mm}
\section{Related works}
\label{sec_relatedwork}
In this section, the works related to the proposed method are briefly reviewed including supervised skeleton based action recognition and unsupervised skeleton based action recognition.

\subsection{Supervised Skeleton based Action Recognition}
\subsubsection {Hand-Crafted Feature based Method} The hand-crafted skeleton features are widely used in early action recognition ~\citep{Weng2017SpatioTemporalNN,Xia2012ViewIH,Vemulapalli2014HumanAR,Evangelidis2014SkeletalQH,Wang2014MiningMF}. For example, Weng et al. \citep{Weng2017SpatioTemporalNN} used Spatio-Temporal Naive-Bayes Nearest-Neighbor to capture spatio-temporal structure of skeleton joints. However, the generalization ability of hand-crafted skeleton features is weak and these methods perform worse on large datasets such as NTU-RGB+D-60~\citep{shahroudy2016ntu}.

\subsubsection {Deep Learning based Method} Depending on the type of network, it can be generally classified
into three categories: CNNs based, RNNs based and GCNs based. In the category of CNNs based methods~\citep{Ke2017SkeletonNetMD,Ke2017ANR,Li2017JointDM,Caetano2019SkeletonIR,Hou2018SkeletonOS,Li2019MultiviewBased3A,Xu2018EnsembleOC,Banerjee2021FuzzyIC,Xia2022LAGANetLA,Zhu2020ACC,Cao2019SkeletonBasedAR}, skeleton sequence is mapped into a color image and fed into CNNs to recognize action classes. For example,  Hou et al.~\citep{Hou2018SkeletonOS} drew skeleton joints with different colors to generate skeleton optical spectra image. Banerjee et al.~\citep{Banerjee2021FuzzyIC} used distance feature, distance velocity feature, angle feature and angle velocity feature to obtain four grayscale images. The fuzzy combination is used to fuse scores extracted from four  grayscale images. Xia et al.~\citep{Xia2022LAGANetLA} utilized convolutions with attention mechanisms to generate local-and-global attention network. Zhu et al.~\citep{Zhu2020ACC}  designed a cuboid CNN model with attention mechanism for skeleton-based action recognition, where a cuboid arranging strategy is used to organize new action representation between all body joints. Although the temporal information is explored to some extent by coding the temporal change into an image, its representation capability in temporal modelling is still relatively limited.

The second category is to treat skeleton as a sequence and use RNNs to extract spatial-temporal information. It focuses more on the temporal information while the spatial information of skeleton joint is not fully explored. To enhance the capturing of spatial information, many methods~\citep{Li2018IndependentlyRN,Li2019DeepIR, Liu2018SkeletonBasedAR,Song2017AnES,Jiang2020ActionRS, Zhang2018FusingGF, Ng2022MultiLocalizedSA} have been proposed.  For example, Jiang et al. \citep{Jiang2020ActionRS} proposed a denoising sparse long short-term memory network to decrease the intra-class diversity and extract more spatial-temporal information. Ng et al.~\citep{Ng2022MultiLocalizedSA} proposed the multi-localized sensitive autoencoder-attention-LSTM to reduce negative  variations such as performers and viewpoints and improve performance. Zhang et al.~\citep{Zhang2018FusingGF} selected a set of simple geometric features to feed into a multi-stream LSTM architecture with a new smoothed score fusion technique to improve recognition accuracy.  However, these approaches cannot effectively capture relationship among joints.

In order to solve this problem, the third approach  uses GCNs~\citep{yan2018spatial,Shi2019TwoStreamAG,Ye2020DynamicGC,Zhang2020SemanticsGuidedNN, Shi2019SkeletonBasedAR,Kong2022MTTMT, Gao2021SkeletonBasedAR, Liu2022SpatialFA, Peng2021SpatialTG,Song2020RichlyAG,Wu2021Graph2NetPG,Liu2021AMG} to capture topological graph structure of skeleton. For example, Yan et al.~\citep{yan2018spatial} proposed spatial-temporal graph convolutional networks (ST-GCN) to extract topological spatial-temporal features, where a static graph is used to capture relationship among joints. However, a static graph is not suitable for all different actions and cannot extract dynamic features among spatial joints. In order to solve this problem, existing methods adaptively learned the topology of skeleton through attention or other similar mechanisms. For example, Liu et al.~\citep{Liu2021AMG} proposed a Graph Convolutional Networks-Hidden conditional Random Field (GCN-HCRF) model to construct multi-stream framework. Generally, GCNs based methods have achieved the state-of-art performance for supervised skeleton based action recognition.

While the supervised learning based methods have greatly advanced in the last few years and achieved good performance, these methods require massive labels for training and cannot effectively work for unlabeled skeleton data. Therefore, unsupervised skeleton based action recognition methods are highly desired.

\subsection{Unsupervised Skeleton based Action Recognition}

\subsubsection{Self-supervised learning based Method} Pretext tasks are designed to extract discriminative features in self-supervised learning based methods. Zheng et al.~\citep{Zheng2018UnsupervisedRL} used the encoder-decoder model and the Generative Adversarial Network (GAN) to reconstruct the skeleton sequence. Su et al.~\citep{Su2020PREDICTC} designed an autoencoder structure with a weak decoder using recurrent neural network to learn more robust features from skeleton sequence. Lin et al.~\citep{Lin2020MS2LMS} proposed two pretext tasks including motion prediction and Jigsaw puzzle recognition to learn  more general representations. However these methods usually used RNNs to extract temporal features where the spatial information are not mined effectively. 

\subsubsection{Contrastive Learning based Method}
Rao et al.~\citep{Rao2021AugmentedSB} used momentum LSTM with a dynamic updated memory bank as the model, and augmented instances of the skeleton sequence are contrasted to learn feature representation. Li et al.~\citep{Li20213DHA} designed a cross-view contrastive learning scheme and leveraged multi-view complementary supervision signal. Thoker et al.~\citep{Thoker2021SkeletonContrastive3A} designed several
skeleton-specific spatial and temporal augmentations to construct skeleton intra-inter contrastive
learning. Lin et al.~\citep{Lin2023ActionletDependentCL} proposed a new actionlet dependent contrastive learning  by treating motion and static regions differently.
Zeng et al.~\citep{Zeng2023Contrastive3H} proposed a Cross Momentum Contrast (CrossMoCo) framework to learn local and global semantic features and used two independent negative	memory banks to improve high-quality of negative samples. Gao et al.~\citep{Gao2023EfficientSC} proposed spatio-temporal contrastive learning using different spatio-temporal observation scenes to build contrastive proxy tasks. Shah et al.~\citep{10203218} proposed Hallucinate Latent Positives for contrastive learning to generate new positives and improve performance. These methods need to design  different positive and negative pairs to improve performance. Zhang et al.~\citep{Zhang2022UnsupervisedSA} proposed a skeleton-based relation consistency learning scheme to expand the contrastive objects from individual instance to the relation distribution between instances, and target at pursuing the relationship consistency learning between different instances. Zhang et al.~\citep{Zhang2022SkeletalTU} used Barlow Twins’ objective function to minimize the redundancy and keep similarity of different skeleton augmentations. However, these methods cannot effectively capture robust and discriminative features for action recognition. Moreover, both the self-supervised learning based methods and contrastive learning based methods suffer from severe overfitting problem and only very small networks can be used, leading to reduced representation capability.    

%\subsection{Overfitting on deep learning}
%{\color{red}Overfitting is a fundamental issue in  deep learning, which cannot optimize model well from observed data to unseen data. In order to reduce model overfitting, many regularization algorithms have been proposed and it includes three main categories from input data, network and label. The first approach is data augmentation~\citep{Shorten2019ASO} to add training dataset and improve the generalization of model such as RandomErasing~\citep{zhong2020random}, Fast autoaugment~\citep{Lim2019FastA}. The second one is regularization based on network such as weight decay~\citep{Loshchilov2017DecoupledWD} and dropout~\citep{Lu2021LocalDropAH}. The third approach is label regularization such as label smoothing~\citep{Li2020RegularizationVS} and label transformation~\citep{Yoo2020RethinkingDA}. } 

\begin{figure*}[ht]
	\centering
	\subfigure[	P\&C~\citep{Su2020PREDICTC}]{\includegraphics[width=0.325\textwidth,height=0.335\textwidth]{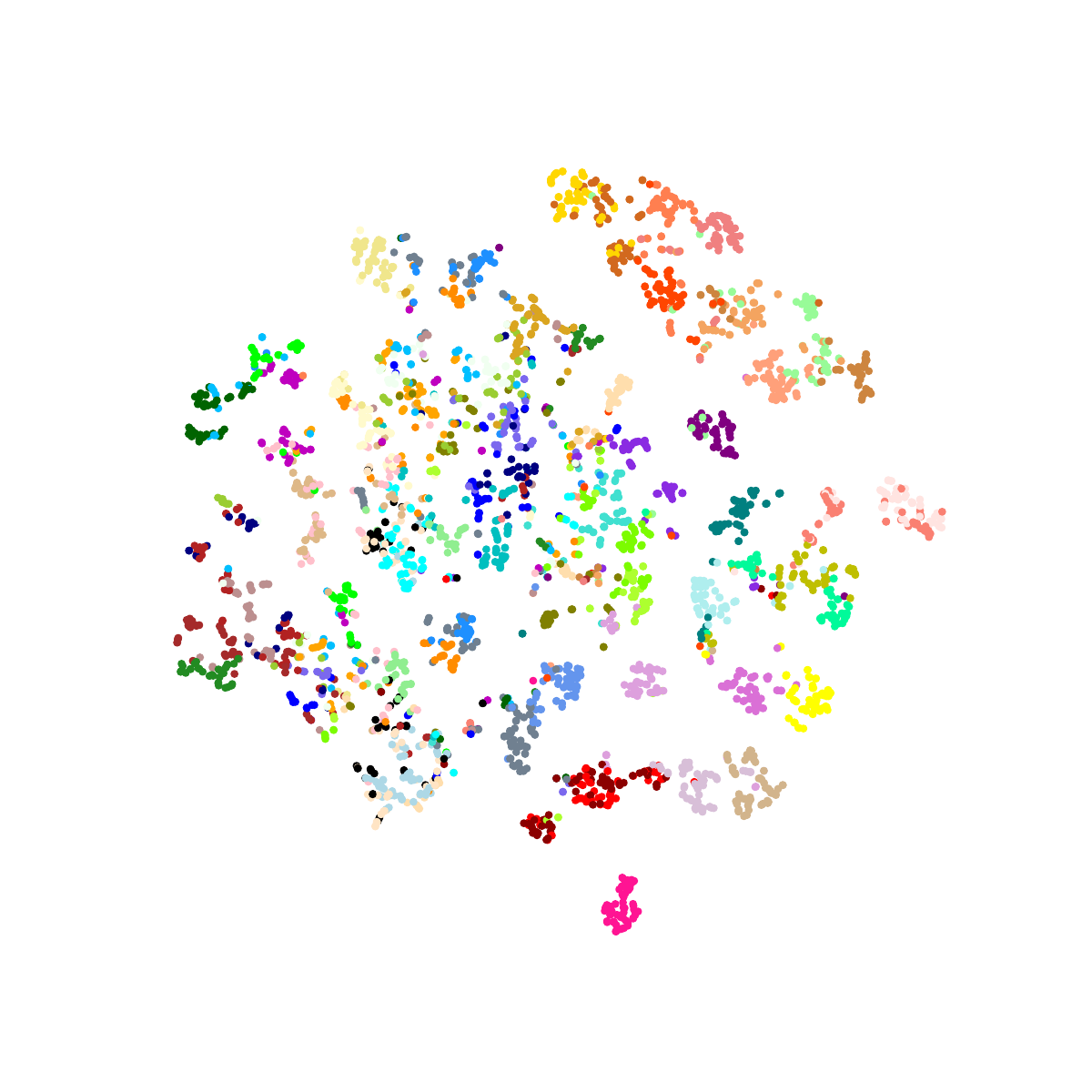}\label{tsnecomp_pc}}
	\subfigure[ASCAL~\citep{Rao2021AugmentedSB}]{\includegraphics[width=0.325\textwidth,height=0.335\textwidth]{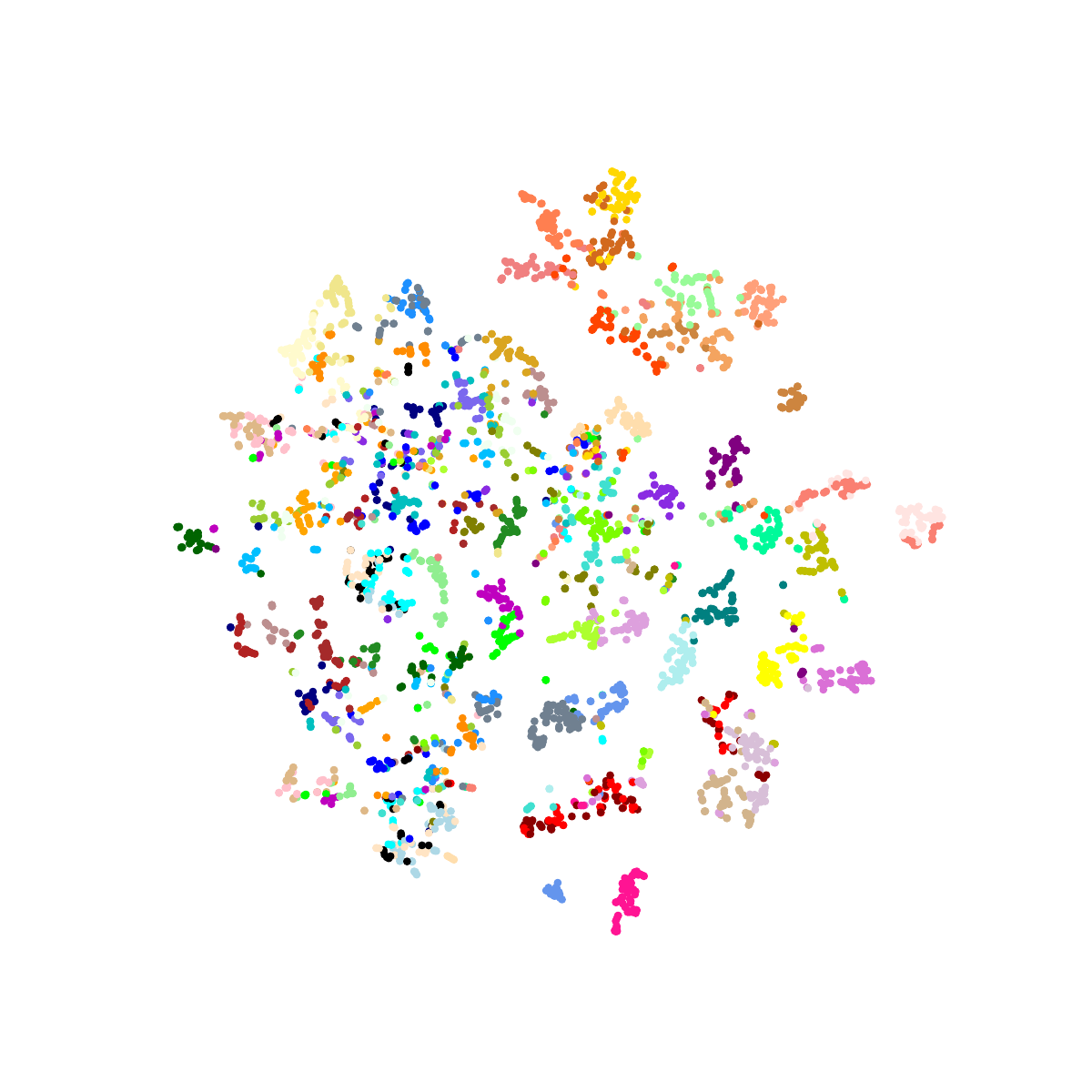}\label{tsnecomp_ascal}}
	\subfigure[Adaptive GCN~\citep{Shi2019TwoStreamAG} with unsupervised learning]{\includegraphics[width=0.325\textwidth,height=0.335\textwidth]{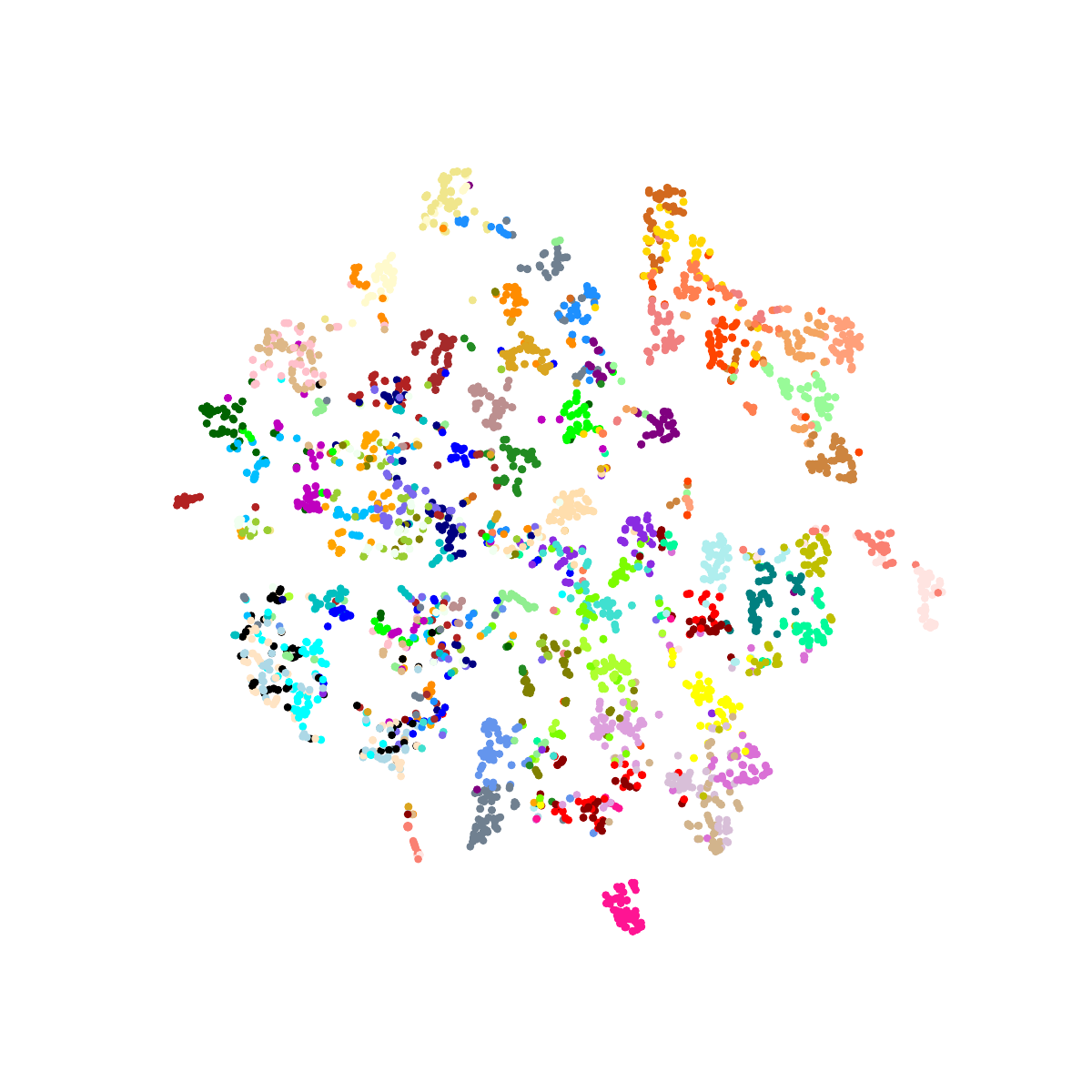}\label{tsnecomp_unsupgcn}}
	
	%\vspace{1cm}
	\subfigure[Proposed U-FEFP]{\includegraphics[width=0.34\textwidth,height=0.35\textwidth]{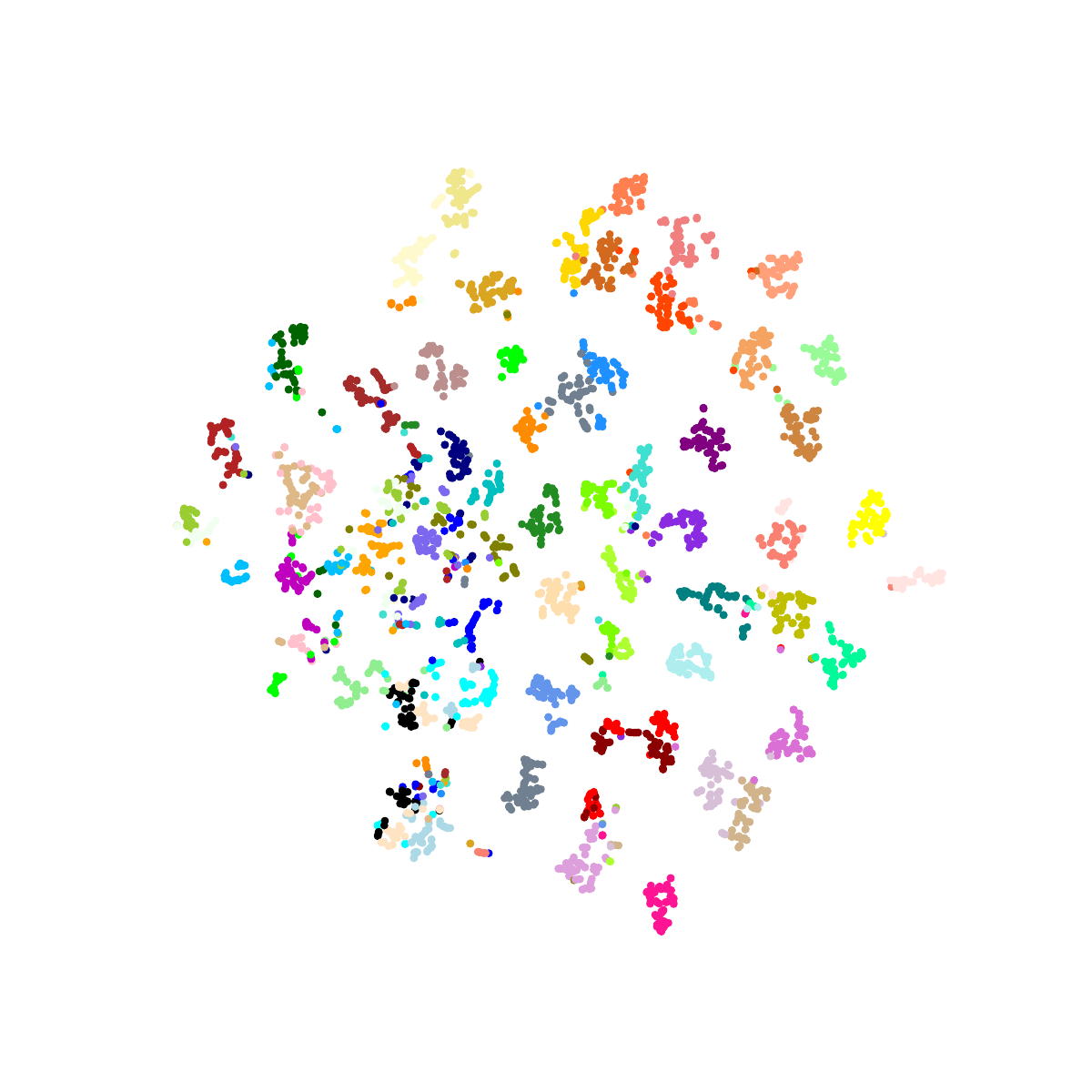}\label{tsnecomp_ust}}
	\subfigure[Adaptive GCN~\citep{Shi2019TwoStreamAG} with supervised learning ]{\includegraphics[width=0.35\textwidth,height=0.34\textwidth]{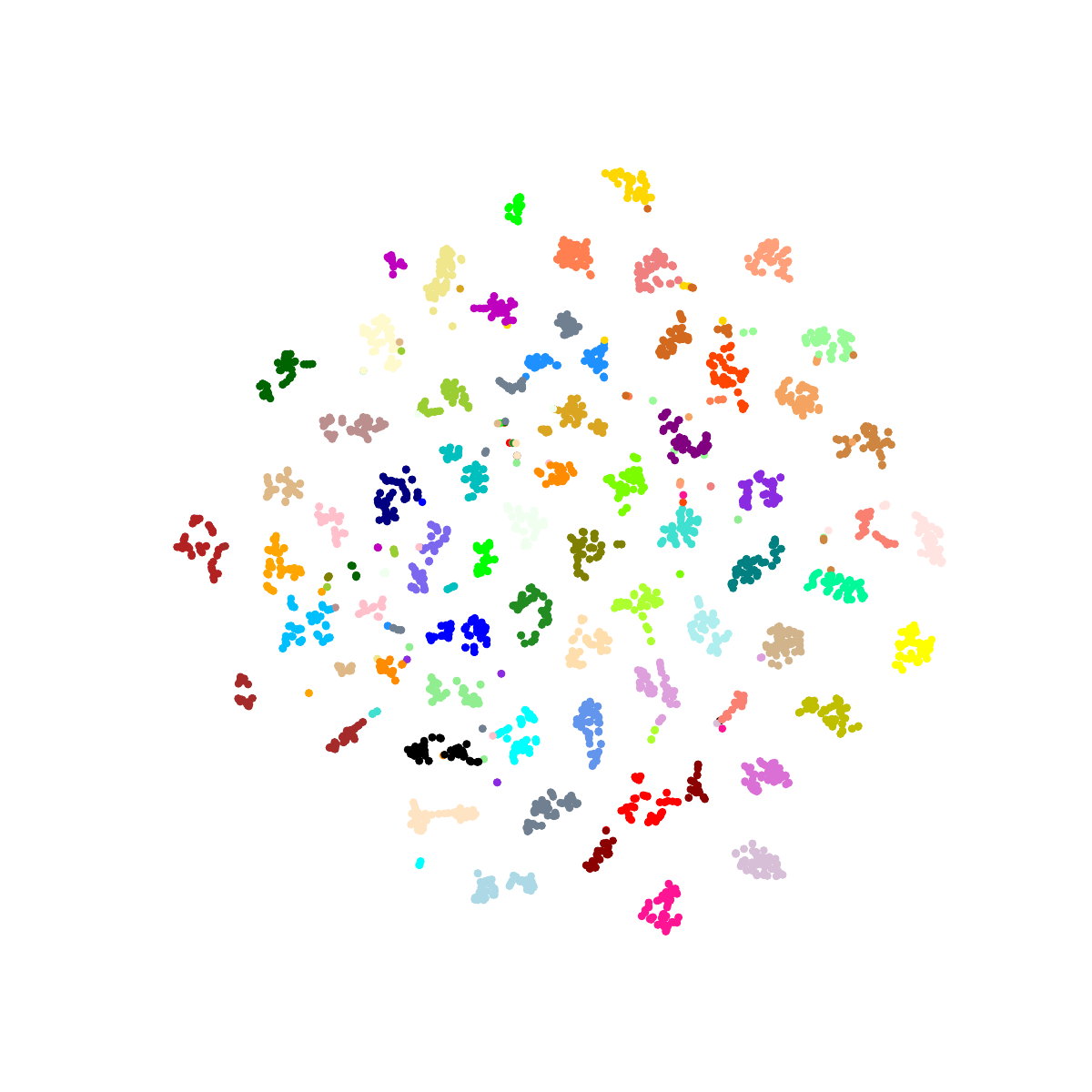}\label{tsnecomp_supgcn} } 
	\caption{t-SNE visualization of the learned features of different methods on the cross-subject of NTU-RGB+D-60. 60 samples are selected for each class on the dataset.  (a) Unsupervised learning based on pretext task, P\&C~\citep{Su2020PREDICTC}. (b) Unsupervised contrastive learning with the momentum LSTM, ASCAL~\citep{Rao2021AugmentedSB}. (c) Unsupervised contrastive learning with the adaptive GCN~\citep{Shi2019TwoStreamAG}. (d) Proposed U-FEFP. (e) Supervised learning with the adaptive GCN~\citep{Shi2019TwoStreamAG}.  }\label{tsnecomp}  
\end{figure*}

\vspace{3mm}
\section{Motivation}
\label{sec_motivation}
As mentioned in the Introduction and Related Work sections, existing unsupervised learning for skeleton based action recognition methods suffer from severe overfitting problem. As shown in Fig. \ref{loss}, the test performance is much worse than the training performance when using the existing models. Detailed descriptions on the experimental setup and analysis are shown in the following Subsection \ref{sec_abl}. Considering that unsupervised learning naturally can use much more data without label than the supervised one, this work focuses more from the perspective of constructing a new model less prone to overfitting  instead of data augmentation and regulation~\citep{Shorten2019ASO, Li2020RegularizationVS,zhong2020random, Lim2019FastA, Yoo2020RethinkingDA,Loshchilov2017DecoupledWD, Lu2021LocalDropAH}. Specifically, we study why one model working well in supervised learning leads to overfitting in unsupervised learning for skeleton based action recognition, and this behaviour has not occurred in the image related unsupervised learning. Instead of directly reducing the number of parameters used in the network \citep{Zhang2022UnsupervisedSA,Zhang2022SkeletalTU} which in turn reduces the capability of the network, this paper first investigates the mechanism behind this severe overfitting problem. Then based on the observation and this overfitting mechanism, our learning framework, U-FEFP, is proposed. 

First, to illustrate the differences between the features learned with unsupervised learning and supervised learning, t-SNE~\citep{Maaten2008VisualizingDU} is used to visualize their embedding clustering. The visual illustration shows how the embedding of the same class of actions form clusters while different classes of actions are separated. Features from three unsupervised learning methods, including unsupervised learning based on pretext task (P\&C~\citep{Su2020PREDICTC}), unsupervised contrastive learning with the momentum LSTM (ASCAL~\citep{Rao2021AugmentedSB}), unsupervised contrastive learning with the adaptive GCN~\citep{Shi2019TwoStreamAG}, are illustrated, and features from the supervised learning with the adaptive GCN~\citep{Shi2019TwoStreamAG} are used for comparison. The t-SNE comparison is shown in Fig. \ref{tsnecomp} using 60 samples from each action class. First, by visualizing the t-SNE illustration of the supervised GCN in Fig. \ref{tsnecomp_supgcn}, it can be seen that the samples are clustered well according to different actions, leading to the good classification results with supervised learning. On the contrary, the t-SNE illustrations of the unsupervised learning with different methods in Figs. \ref{tsnecomp_pc}, \ref{tsnecomp_ascal} and \ref{tsnecomp_unsupgcn} show that the samples are also grouped to some extent, but not according to their action classes, thus producing poor results. Especially comparing the t-SNE illustrations of the adaptive GCN under supervised and unsupervised learning in Figs. \ref{tsnecomp_supgcn} and \ref{tsnecomp_unsupgcn}, it can be clearly seen that with the same network, the features are learned completely differently, in terms of their clustering behaviour to the action classes. The features of the supervised learning GCN are grouped according to the action classes while the features with the unsupervised learning GCN are also grouped to some extent, but not strongly related to the action classes. Intuitively, this can be analyzed as the choice of the negative samples not highly related to  action recognition, leading to the difficulty in choosing negative samples in unsupervised learning. Moreover, considering the samples are also grouped to some extent in unsupervised learning, this demonstrates that the skeletons are also high-level features that can be discriminated easier than action classes.

As a matter of fact, the skeleton sequences are already relatively high-level and low-dimension representations. In such a case, unsupervised learning tends to produce features that directly discriminate or reconstruct samples, and such features may not be useful for action recognition. In other words, the features learned in the unsupervised way distribute in a high-level manifold that is not aligned with the high-level feature manifold of the action recognition. Accordingly, the loss in the unsupervised learning can be very small while the loss of the action recognition is very high, leading to the overfitting problem. To overcome this problem, we propose to generate \textit{rich distributed spatial-temporal} features \textit{containing all information of the original skeleton} in unsupervised learning. The \textit{rich distributed spatial-temporal} features contain distributed features that can be useful for action recognition, instead of pushing features to discriminate certain samples that may narrow the representation capability of the features. Constraining the features to be \textit{containing all information of the original skeleton} encourages the network to preserve all useful information. To the best of our knowledge, this is the first research that clearly points out to learn such features in the  unsupervised skeleton based action recognition. Based on this observation, we propose a U-FEFP learning framework, which is explained in the next section. The t-SNE illustration of our U-FEFP learning framework is shown in Fig. \ref{tsnecomp_ust}. Compared to the other unsupervised learning methods shown in Figs. \ref{tsnecomp_pc}, \ref{tsnecomp_ascal} and \ref{tsnecomp_unsupgcn}, our U-FEFP clearly produces features that are better aligned with the action classes. Although certain samples may deviate from the action centers, they are also away from other action centers, making them easier for recognition.

\begin{figure*}[ht]
	\centering
	\includegraphics[width=0.92\textwidth,height=0.52\textwidth]{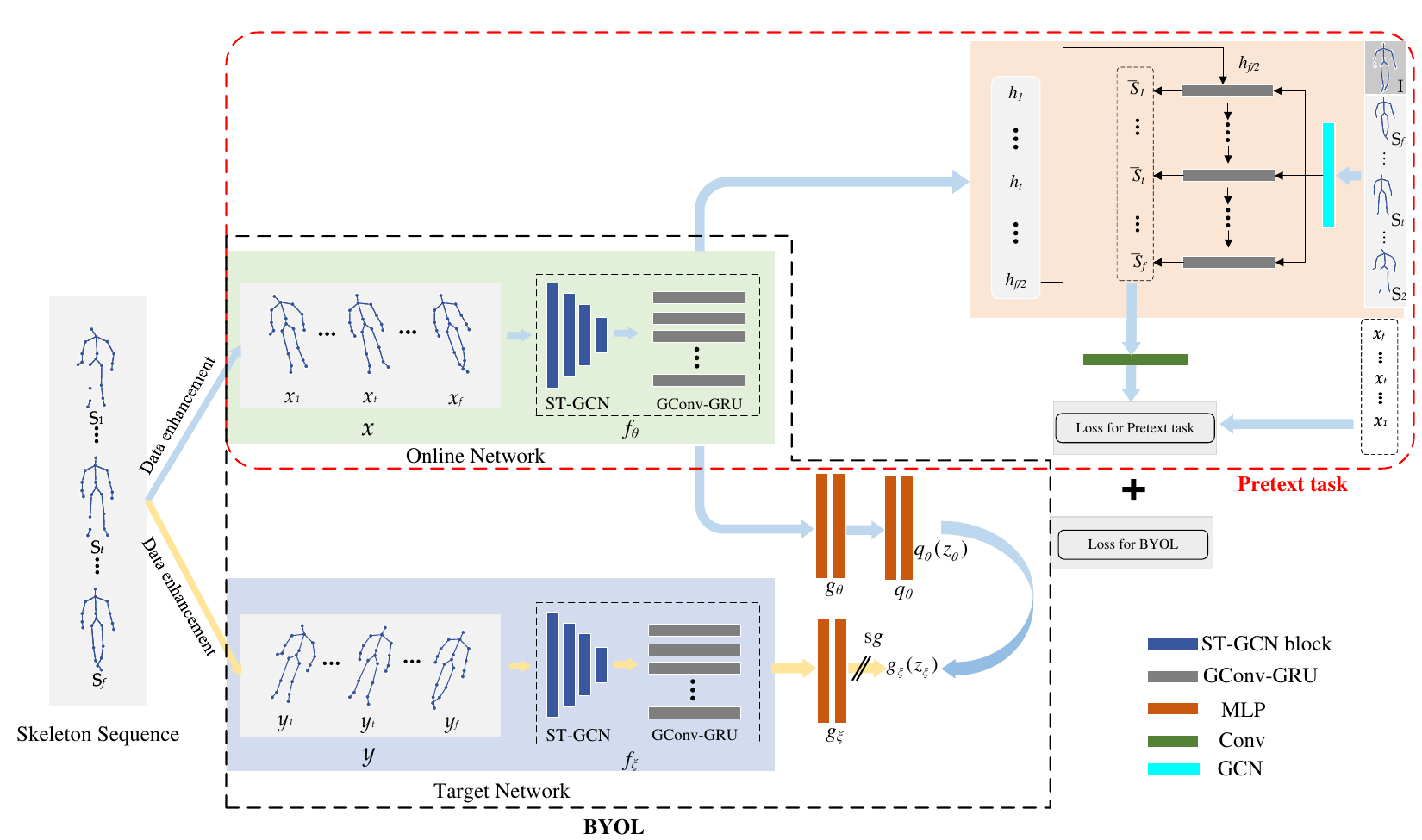}
	\caption{The framework of the proposed U-FEFP, with unsupervised BYOL based feature enrichment learning and unsupervised pretext task based fidelity preservation learning. It consists of an online network (in green), a target network (in blue) and a reversed prediction network (in beige). The online network is trained to learn rich representations and the target network is slowly updated by the exponential moving average of the online network to make them asynchronous. The reversed prediction network is used to reconstruct the skeleton sequence with the features generated by the online network. The BYOL based contrastive learning (within the black dash box) and the reversed prediction (pretext task) based learning (within the red dash box) are used to keep similarity of different skeleton augmentations at feature and instance level, respectively.}\label{framework}  
\end{figure*}

\section{Proposed Method}
\label{sec_proposed}

The framework of the proposed unsupervised spatial-temporal feature enrichment and fidelity preservation network (U-FEFP) is shown in Fig.~\ref{framework}, consisting of two parts: unsupervised BYOL based feature enrichment learning and unsupervised pretext task based fidelity preservation learning. The unsupervised BYOL based feature enrichment learning, including the online network and target network, is developed for spatial-temporal feature transformation, in order to obtain a rich distributed spatial-temporal feature representation. The unsupervised pretext task based fidelity preservation learning, including the online network and the decoding network, is developed for spatial-temporal feature fidelity preservation, in order to keep the original skeleton information. The details of the proposed U-FEFP are presented in the following.

\subsection{Unsupervised BYOL based Learning for Spatial-temporal Feature Enrichment}
\subsubsection{Spatial-temporal Feature Transformation Network}
A spatial-temporal graph convolution subnetwork (ST-GCN) followed by a graph convolutional GRU network (GConv-GRU) is developed as the basic architecture of the online network and target network used in our unsupervised BYOL based learning, to produce the spatial-temporal features. ST-GCN takes advantage of the graph convolution to extract the spatial-temporal feature of the skeleton. With the expressive power of graph convolution in processing non-grid data like skeleton, it can obtain rich spatial features. Moreover, considering the temporal change of each joint among frames is also important in characterizing the spatial features of a skeleton to be representative and discriminative against others, ST-GCN is used to obtain spatial and short-term temporal features. In each ST-GCN block, it consists of one spatial graph convolution extracting the spatial information and one temporal convolution mining the short temporal information. The basic structure of the graph convolution is the same as \citep{yan2018spatial}, which is not further detailed here. Four ST-GCN blocks are used and the numbers of convolution kernels are 32, 64, 128 and 512, respectively. In order to reduce the computation, the stride of temporal convolution is set to 2 in the fourth ST-GCN block, which halves the length of the temporal features.

For capturing the long-term temporal features, a GConv-GRU network is used, which aggregates the spatial and short-term temporal features from ST-GCN in order to achieve long-term information. On one hand, it is found that using ST-GCN for  complete spatial-temporal feature extraction is easily overfitting, since multiple ST-GCN layers are needed to extract the long-term temporal features, making the network complex. By contrast, in our method, as shown in Fig.~\ref{framework}, only four ST-GCN blocks (versus ten blocks in the conventional ST-GCN methods \citep{yan2018spatial}) are used to reduce overfitting. On the other hand, GConv-GRU, due to its recurrent structure, is more suitable for decoding features in the following temporal pretext task (described in the next subsection). Therefore, a sequential architecture combining the ST-GCN and GConv-GRU is used in this paper.

\begin{figure}[t]
	\centering
	\includegraphics[width=0.48\textwidth,height=0.3\textwidth]{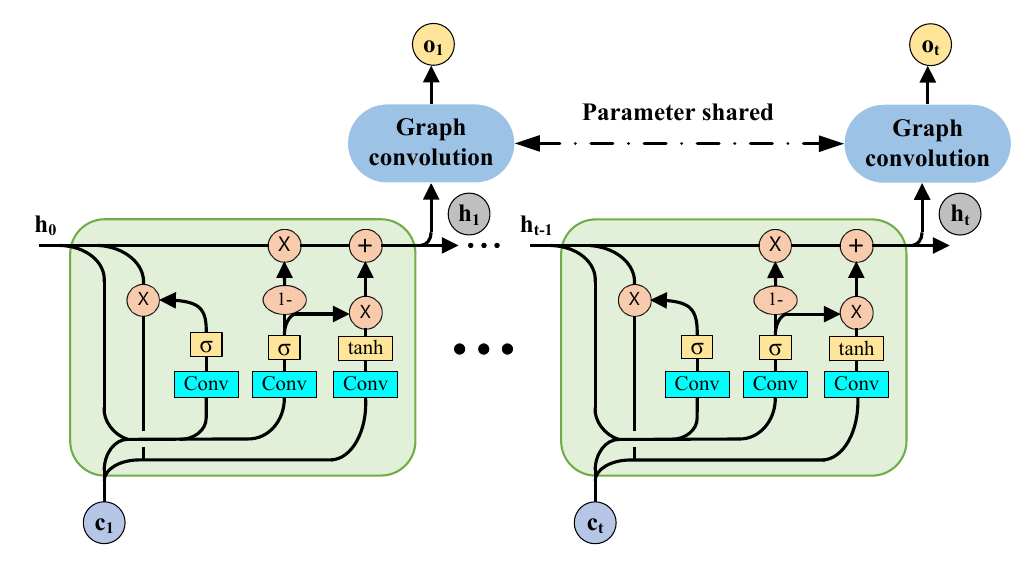}
	\caption{The structure of the GConv-GRU }\label{gru}  
\end{figure}

For the GConv-GRU, while the recurrent structure aggregates the temporal information, its sequential processing also incurs great computation if the processing of each step is computationally expensive. Here, considering the features are captured via the ST-GCN with graph convolution, the spatial structure information is already contained in the input features to the GConv-GRU. Therefore, in order to reduce computation, general convolution, with 1*1 kernel for per-joint processing, is used in the recurrent update of each GRU step. To make the input processing consistent with the recurrent processing, general convolution is also used in the input processing.  The hidden output of the GConv-GRU is further enhanced with graph convolution, which can be computed in parallel over all time steps with less computation complexity and enhancing the temporal features with the spatial structure. The update of the GConv-GRU is shown in Fig.~\ref{gru}.
%and expressed as follows.

%It is worth noting again that the input is already processed with the ST-GCN to contain the spatial structure information.

%\begin{equation}
%\bf{{z_t}} = \sigma ({\bf{W_z}}*{c_t} + {U_z}*{h_{t - 1}}+b_{z})
%\end{equation}
%\begin{equation}
%\bf{{r_t}} = \sigma ({W_r}*{c_t} + {U_r}*{h_{t - 1}} + {b_r})
%\end{equation}
%\begin{equation}
%\bf{{\mathop {h{_t}}\limits^ \sim}} = \textit{tanh} ({W_h}*{c_t} + {U_h}*({r_t} \circ {h_{t - 1}}) + {b_h})
%%\end{equation}
%\begin{equation}
%\bf{{h_t}} = (1 - {z_t}) \circ {h_{t - 1}} + {z_t} \circ {\mathop {h_{t}}\limits^ \sim}
%\end{equation}
%\begin{equation}
%\bf{{o_t}} = \textit{relu}(W_o * {h_t} \cdot A +b_o)
%\end{equation}
%where $*$ is the general convolution operation. $\sigma$, $tanh$, and $relu $ are activation function. $\bf{c_{t}}$ is output of the ST-GCN subnetwork, and $\bf{h_t}$ is the hidden output. $\bf{o_t}$ is the final output produced with the graph convolution processing of the hidden output $\bf{W_o * {h_t} \cdot A}$, where $\bf{A}$ represents the graph of a skeleton. The number of convolution kernels is set to 256 in the GConv-GRU. This spatial-temporal feature transformation network architecture is used for both the online network and target network.
% Followed by the GConv-GRU, a global pooling is used to generate the final 256 dimension feature.

\subsubsection{BYOL based Feature Enrichment Learning}
As mentioned in the Motivation, for unsupervised learning, rich distributed features are highly desired. It is required to produce a rich set of distributed high-level representation features that is useful to discriminate different samples and useful for the downstream high-level task, i.e., action recognition. Naively we can generate a set of features with a network of random parameters. However, this cannot provide view-invariant (shift-invariant, pose-invariant, etc.) high-level features. Thanks to the BYOL \citep{Grill2020BootstrapYO} based feature learning, rich distributed features can be learned with two asymmetric networks, i.e., an online network and a target network as shown in Fig.~\ref{framework}, by pulling together the features of different augmented versions of one sample.

As shown in the Fig.~\ref{framework}, data augmentation is first used to enhance a skeleton sequence to different views. Suppose that an original skeleton sequence $\rm{\bf{S}} = ({\bf{S}_1}, \cdot  \cdot  \cdot ,{\bf{S}_f})$ contains
$f$ consecutive skeleton frames, where $\bf{S}_{i}  \in \mathbb{R}^{N \times 3} $ is 3D coordinates of N skeleton joints. The data augmentation strategy in~\citep{Thoker2021SkeletonContrastive3A} (i.e., spatial augmentation and temporal augmentation) and rotation are  used to transform $\bf{S}$ into its augmented versions $\bf{x}$ and $\bf{y}$. For spatial augmentation, pose augmentation and joint jittering are randomly selected. For temporal augmentation, temporal crop-resize by randomly selecting a starting frame and a sub-sequence length, and then resizing the sub-sequence to a fixed length is used. For rotation, an random axis with random rotation is selected. The augmented skeleton sequence does not change the graph structure of a skeleton and thus same graph convolution can be used. 

%Some contrastive learning based action recognition methods~\citep{Rao2021AugmentedSB,Li20213DHA,Thoker2021SkeletonContrastive3A} utilize the large amount of negative samples requiring much memory-bank. In this letter, we follow BYOL~\citep{Grill2020BootstrapYO} scheme to build the contrastive learning architecture as shown in Fig.~\ref{framework}. 
The two different views of the samples are then processed by the online network and target network, generating the spatial-temporal features.
Then two nonlinear projectors ${g_\theta}$  and ${g_\xi}$ are used  to project the hidden features into a new feature space. The two nonlinear projectors use same structure and are updated in the same way with online network and target network. The nonlinear projector consists of two fully connected layers. The first fully connected layer is of 1024 neurons followed by a batch normalization layer and a relu activation layer. The second one is of 512 neurons generating features without the normalization and activation. This up-projects the features back to 512 channels to enrich the representation. 

As in BYOL framework, asymmetric architecture is used and a predictor ${q_\theta} $ using the same network as the nonlinear projector ${g_\theta}$ is added only to the online branch to produce the prediction $\bf{{q_\theta}(z_{\theta}})$, where  $\bf{{z_\theta}} $ is output of the projector ${g_\theta}$. For the target network, the stop-gradient operation is used after  the nonlinear projector $\bf{{g_\xi}} $ and obtains feature $\bf{{g_\xi}(z_{\xi})}$, where  $\bf{{z_\xi}} $ is output of the target network. Then $\bf{{q_\theta}(z_{\theta}}) $ and $\bf{{g_\xi}(z_{\xi})}$ are normalized with the ${\ell _2}$-norm separately as

\begin{equation}
	\bf{{\bar q_\theta }({z_\theta }) \buildrel \Delta \over = {q_\theta }({z_\theta })/{\left\| {{q_\theta }({z_\theta })} \right\|_2}} \label{eq1}
\end{equation}

\begin{equation}
	\bf{{{\bar g}_\xi }({z_\xi }) \buildrel \Delta \over = {g_\xi }({z_\xi })/{\left\| {{g_\xi }({z_\xi })} \right\|_2}	} \label{eq2}
\end{equation}

Finally, Mean Square Error (MSE) objective function between $\bf{\bar{q_\theta}(z_{\theta}}) $ and $\bf{\bar{g_\xi}(z_{\xi})} $ is used to construct the self-supervised loss  and can be expressed as:

\begin{equation}
	{L_{\theta ,\xi }} = \left\| {{{\overline {\bf{q}} }_{\bf{\theta }}}({{\bf{z}}_{\bf{\theta }}}) - {{\overline {\bf{g}} }_{\bf{\xi }}}({{\bf{z}}_{\bf{\xi }}})} \right\|_2^2 \label{eq3}
\end{equation}
which can be further transformed by substituting $\bf{\bar{q_\theta}(z_{\theta}}) $ and $\bf{\bar{g_\xi}(z_{\xi})} $ with Eqs. (\ref{eq1}) and~(\ref{eq2}), respectively, to the following:%The equations~\ref{eq1} and~\ref{eq2} are updated into  equation~\ref{eq3}. The self-supervised loss ${L_{\theta ,\xi }}$ can be obtained as

\begin{equation}
	{L_{\theta ,\xi }} = 2 - 2 \cdot \frac{{\left\langle {{{\bf{q}}_{\bf{\theta }}}({{\bf{z}}_{\bf{\theta }}}),{{\bf{g}}_{\bf{\xi }}}({{\bf{z}}_{\bf{\xi }}})} \right\rangle }}{{{{\left\| {{{\bf{q}}_{\bf{\theta }}}({{\bf{z}}_{\bf{\theta }}})} \right\|}_2} \cdot {{\left\| {{{\bf{g}}_{\bf{\xi }}}({{\bf{z}}_{\bf{\xi }}})} \right\|}_2}}}	
\end{equation}
The loss is symmetrized and a symmetric loss ${L_{\theta ,\xi }}^{'}$ can be obtained by feeding the $\bf{x}$ and $ \bf{y}$ into target network and online network, respectively. Finally, the learning loss is obtained as ${L_{BYOL}} = {L_{\theta ,\xi }} + {L_{\theta ,\xi }}^{'}$.

In the training process, weights $\xi$ of target network are updated using the exponential moving average of the online network weight $\theta$ which follows $\tau \xi  + (1 - \tau )\theta  \to \xi $. This allows the online network and target network to be always asymmetric. The online and target GConv-GRU network produce a temporal feature with half the time steps of the skeleton sequence. While the features of all time steps can be processed as above, in this paper for simplicity, a global pooling over the temporal dimension is used to generate the final 256 dimension feature and then processed.

This BYOL based feature learning enables the online network to generate rich distributed high-level features. Simply speaking (as an extreme case for intuitive understanding), the target network produces a rich set of randomly combined features, while the learning updates the online network and target network to produce view-invariant high-level features. The asymmetric updating avoids them to generate the trivial solutions of zero or other fixed representations. Therefore, the online network produces rich distributed high-level features, which is further constrained by the pretext task (described in the following  subsection) and  used for the final action recognition.

\begin{figure}[tb]
	\centering
	\includegraphics[width=0.5\textwidth,height=0.34\textwidth]{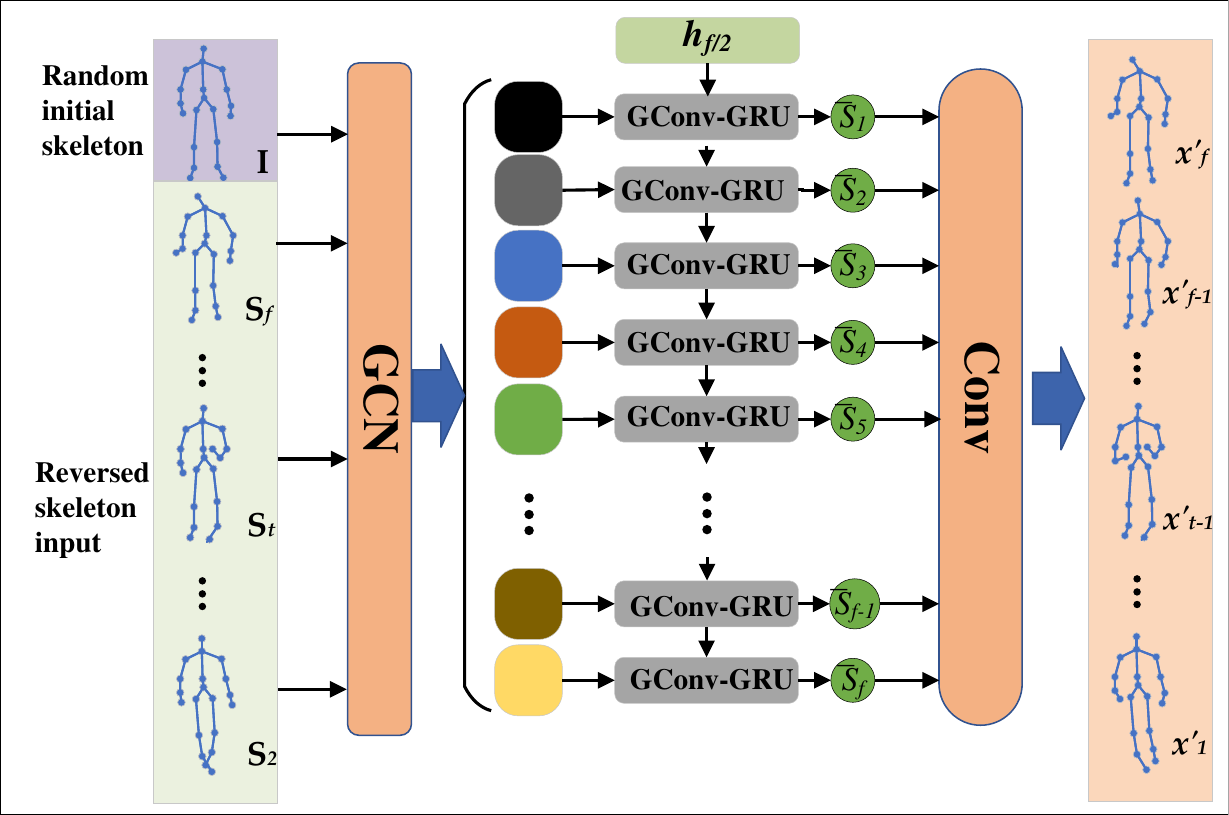}
	\caption{The structure of the decoder in the unsupervised pretext task based learning.}  \label{pretext}   %It consists of three parts: one layer GCN, GConv-GRU and Convolution.
\end{figure}

%\subsection{Pretext Task for Self-supervised Learning}
\subsection{Unsupervised Pretext Task based Learning for Spatial-temporal Feature Fidelity Preservation}

While the above BYOL based learning generates rich distributed features that can keep the similarity of augmented different-view skeleton data at instance level, it cannot ensure the capability of the generated features to be able to classify all actions. In other words, the representation space of the generated features may be reduced since there is no constraint in discriminating different samples or actions.  Therefore, to generate features that are not only rich distributed but also full and contain as much information of the original skeleton as possible, a fidelity preservation constraint is required.

In this paper, an unsupervised pretext task based learning is designed using reversed prediction. Motivated by the pretext task using encoder-decoder network~\citep{Su2020PREDICTC,Zheng2018UnsupervisedRL}, the feature of the online network is used as the encoder feature, and a decoder is used to predict the skeleton sequence. To be specific, the skeleton sequence in the reverse order is used  with each skeletal joint predicted. In this way, the hidden state obtained from the encoder (which is at the time step of the last skeleton) can be directly used for predicting the skeleton at its corresponding time step.
The decoder used in this paper consists of one GCN, one GConv-GRU and one Convolution as shown in Fig.~\ref{pretext}. At each time step, the previous skeleton (also in the reverse order) is first fed into GCN to generate features $\bf{d}$ as input, and with the recurrent hidden feature from GConv-GRU at the previous step, the network predicts the skeleton. For the first time step, the encoder (online network) feature is used as the initial recurrent hidden feature and a randomly initialized skeleton is used as the input. The update process can be expressed as 

\begin{equation}
	\bf{({\mathop h\limits^\_}_t,{\mathop S\limits^\_}_t)} = \left\{ {\begin{array}{*{20}{c}}
			{\Theta (\bf{{h_{f/2}},{d_1}){\rm{    }}}}&{t = 1}\\
			{\Theta ({\bf{{\mathop h\limits^\_ }_{t - 1}}},\bf{{d_{t - 1}}}){\rm{    }}}&{t > 1}
	\end{array}} \right.
\end{equation}

where $\Theta ( \cdot )$ denotes the decoder GConv-GRU. $\bf{{h_{f/2}}}$ is the encoder feature from the online network, i.e., the hidden state at the last time-stamp. When the decoding is initialized (t=1), $\bf{{h_{f/2}}}$ and randomly initialized $\bf{{d_1}}$ is provided to the decoder GConv-GRU, producing the recurrent hidden features $\bf{\mathop h\limits^\_}_1$, and the output feature $\bf{\mathop S\limits^\_}_1$. Then $\bf{\mathop S\limits^\_}_1$ is processed with convolution to produce the first skeleton (in the reverse order). For the following time steps, $\bf{{d_{t-1}}}$, the feature of the skeleton at the previous time step, is used as the input and $\bf{\mathop h\limits^\_}_{t-1}$ is used as the recurrent input. Finally, MSE between the output sequence $\bf{x'}$ and $\bf{\mathop x\limits^\_}$ (reversed sequence of $\bf{x}$)  is used as loss of the pretext task based unsupervised learning, ${L_P}=\left\|\bf{x'}- \bf{\mathop x\limits^\_} \right\|_2^2$. The number of convolutional kernel is the same with spatial GCN of the fourth ST-GCN block in the online network.

This unsupervised pretext task based learning enforces the features generated from the online network to contain all the information of the skeleton so as to predict the original skeleton. Together with the BYOL based learning, the proposed U-FEFP learning framework is jointly trained by using the total loss $L = {L_{BYOL}} + {L_P}$, to generate rich distributed and fidelity preserved features. The designed spatial-temporal feature transformation network of combining ST-GCN and GConv-GRU makes the features rich of spatial-temporal information useful for action recognition.

\section{Experimental Results}
\label{sec_exp}
\subsection{Datasets}
Three widely used datasets, including the NTU RGB+D-60  dataset, NTU-RGB+D-120  dataset and PKU-MMD dataset, are used for evaluating the proposed method.

{\bf{NTU RGB+D-60  dataset (NTU-60)}}: NTU-60~\citep{shahroudy2016ntu} is one of the largest indoor-captured dataset for human action recognition task, which has been currently widely used. It is performed by 40 persons with different ages. This dataset contains 4 million frames and 56880 skeleton sequences captured by the Microsoft Kinect v2, and it consists of two side views, front view and left, right 45 degree views. We adopt the same training and testing protocols including the cross-subject (X-sub) and the cross-view (X-view) settings, as in \citep{shahroudy2016ntu}, which is not further detailed here.%: (1) cross-subject (X-sub): training data comes from 1, 2, 4, 5, 8, 9, 13, 14, 15, 16, 17, 18, 19, 25, 27, 28, 31, 34, 35, 38 subjects and the remaining samples are as testing data.  (2) cross-view (X-view): samples of camera views  2 and 3 are reserved for training and testing data comes from camera view 1.

{\bf{NTU-RGB+D-120  dataset (NTU-120)}}: NTU-120~\citep{Liu2020NTUR1} is an extended version of NTU-60. It consists of 114480 action clips that are captured from 106 distinct human subjects. The action samples are captured from 155 different camera viewpoints. The subjects in this dataset are in a
wide range of age distribution (from 10 to 57) and from different cultural backgrounds (15 countries), which brings very realistic variation to the quality of actions. We use cross-subject (X-Sub) and cross-setup (X-Set) adopted in~\citep{Liu2020NTUR1} to evaluate the proposed method.

{\bf{PKU-MMD dataset}}: PKU-MMD dataset~\citep{Liu2020ABD}  has nearly 20000 action clips in 51 action categories. Two subsets PKU-MMD
I and PKU-MMD II are used in the experiments. PKU-MMD II is more challenging than PKU-MMD I due to higher level of noise. Experiments are conducted on the cross subject
(X-Sub) benchmark for both subsets.

\subsection{ Implementation Details }
{\bf{Unsupervised Pre-training}}: We use the PyTorch framework to implement the proposed U-FEFP and run it on four Tesla A100 GPUs. LARS~\citep{Zhang2022UnsupervisedSA} is selected as optimizer and trained for 1500 epochs with a cosine decay schedule. The learning
rate starts at 0 and is linearly increased to 2.0 in the first 25 epochs of training and then decreased to 0.001 by a cosine
decay schedule. We follow the paper~\citep{Li20213DHA} to downsample 50 frames for each skeleton sequence. Target decay rate $\tau$ used in the BYOL based learning is set to 0.99, which is verified in the following ablation study.

{\bf{Linear Evaluation Protocol}}: The online network is frozen, and a fully connected layer is appended to online network and trained for action recognition task. Cross Entropy loss of action recognition is used as the objective function.

\begin{table}[tb]
	\setlength{\belowcaptionskip}{2pt}
	\setlength{\tabcolsep}{2pt}
	\begin{center}
		\caption{Comparison of different feature transformation networks as online network} \label{a}% on the NTU-60
		\begin{tabular}{l  c c } \hline		
			Method& X-Sub (\%) & X-View (\%) \\  \hline 
			MS-G3D~\citep{Liu2020DisentanglingAU}+ BYOL & 50.16& 54.92\\
			CTR-GCN~\citep{Chen2021ChannelwiseTR}+ BYOL & 51.25& 55.68\\ 
			2s-AGCN~\citep{Shi2019TwoStreamAG} + BYOL & 53.68& 56.89\\
			ST-GCN~\citep{yan2018spatial}+ BYOL   &  75.42&  79.50\\
			\hline
			online network v1 + BYOL &  74.88&  80.26\\
			online network v2 + BYOL &  77.85&  82.68\\
			online network v3 + BYOL &  71.22&  76.98\\
			online network v4 + BYOL &  78.12&  82.80\\
			online network v5 + BYOL &  79.31&  84.20\\
			\textbf{Proposed online network + BYOL} &\textbf{80.55}& \textbf{85.62}\\
			%Proposed online network + pretext task &67.23& 73.20\\
			%\textbf{UST-FEFP}  & \textbf{80.21}& \textbf{85.79}\\
			\hline 		
		\end{tabular}
	\end{center}
\end{table}

\begin{table}[tb]
	\setlength{\belowcaptionskip}{2pt}
	\setlength{\tabcolsep}{2pt}
	\begin{center}
		\caption{ Comparison of  different modules in the proposed method } \label{exp_modules}% on NTU RGB+D 60 dataset
		\begin{tabular}{l  c c } \hline		
			Method& X-Sub (\%) & X-View (\%) \\  \hline 
			Proposed online network + BYOL &80.55& 85.62\\
			Proposed online network + pretext task &67.88& 73.96\\
			%ST-GCN~\citep{yan2018spatial} + pretext task &64.94& 70.28\\
			%skeleton  reconstruction network 1 + pretext task &67.12& 72.92\\
			\textbf{U-FEFP}  & \textbf{82.50}& \textbf{87.52}\\
			\hline 		
		\end{tabular}
	\end{center}
\end{table}

\begin{figure*}[thbp!]
	\centering
	\begin{tabular}{@{\extracolsep{\fill}}c@{}c@{\extracolsep{\fill}}}
		\includegraphics[width=0.5\linewidth]{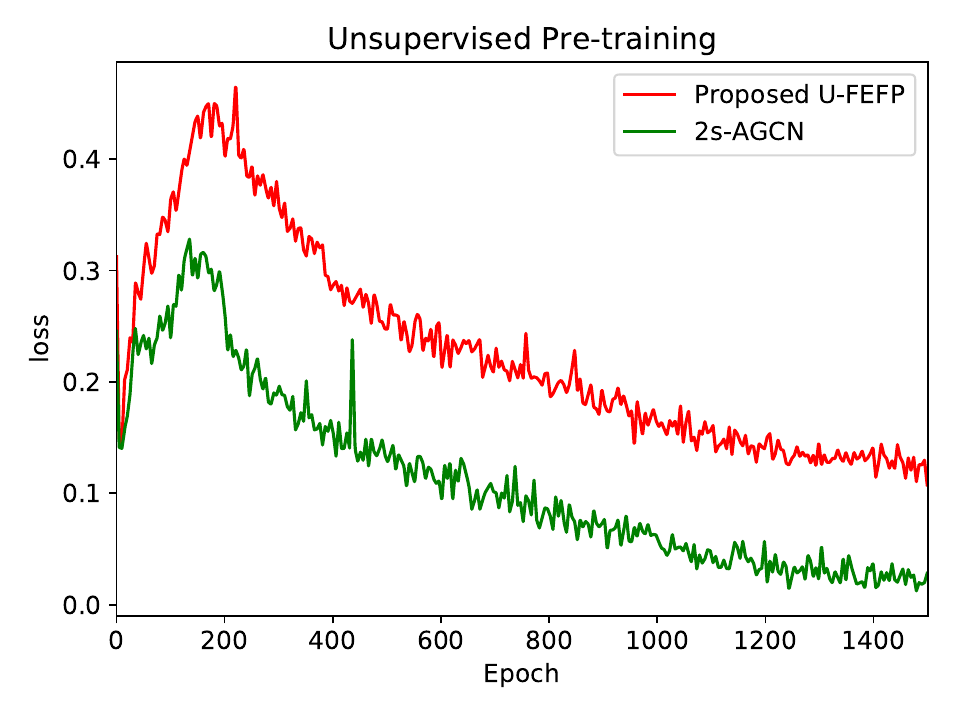} &
		\includegraphics[width=0.5\linewidth]{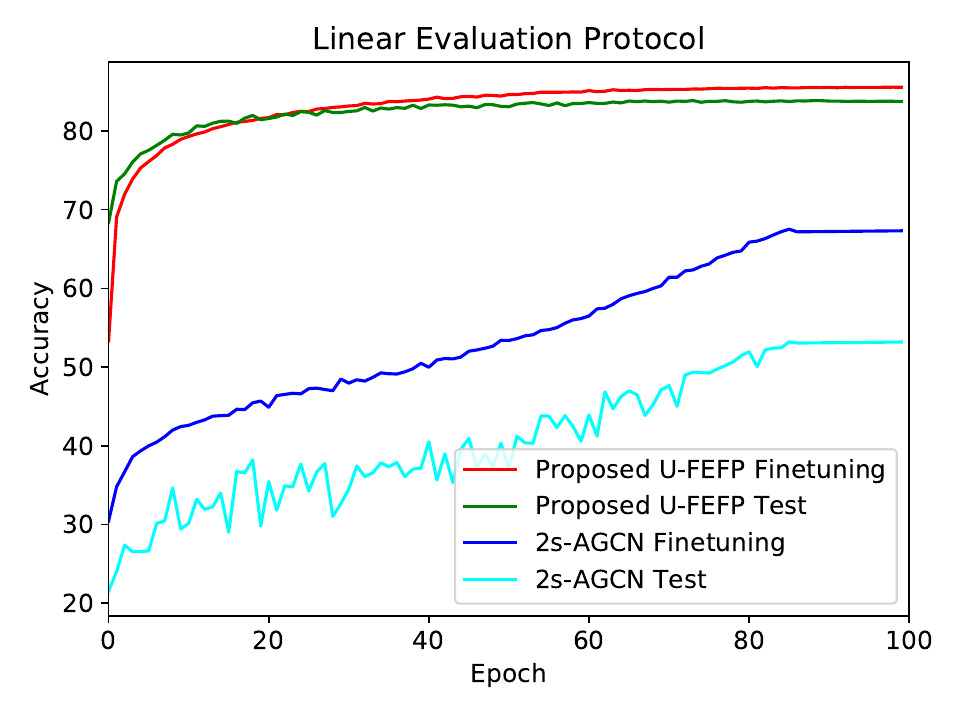}\\
		(a) & (b)\\
	\end{tabular}
	\caption{Comparisons of the proposed U-FEFP and 2s-AGCN  in the process of unsupervised pre-training (a) and fine-tuning/test in the linear evaluation protocol (b), respectively.}
	\label{loss}
\end{figure*}

\subsection{Ablation Study}
\label{sec_abl}
In this section, the effectiveness of our proposed U-FEFP is validated from five aspects: evaluation of the proposed online network, combining the BYOL learning and pretext task based learning, different target decay rates, different batch sizes and semi-supervised learning. The experiments on the NTU-60 dataset are used for all the ablation studies.

{\bf{Evaluation of the proposed online network}}: In order to validate the proposed spatial-temporal feature transformation network as the online network, the BYOL scheme is used with different models as online network for comparison, including MS-G3D~\citep{Liu2020DisentanglingAU}, CTR-GCN~\citep{Chen2021ChannelwiseTR}, 2s-AGCN~\citep{Shi2019TwoStreamAG} and ST-GCN~\citep{yan2018spatial}. The configurations are listed as follows and all configurations are trained from scratch in the same way as the proposed method.

\begin{itemize}
	\item MS-G3D~\citep{Liu2020DisentanglingAU}+ BYOL: MS-G3D~\citep{Liu2020DisentanglingAU} network is used as the online and target networks for the BYOL based learning.
	\item CTR-GCN~\citep{Chen2021ChannelwiseTR}+ BYOL: CTR-GCN~\citep{Chen2021ChannelwiseTR} network is used as the online and target networks for the BYOL based learning.
	\item 2s-AGCN~\citep{Shi2019TwoStreamAG} + BYOL: 2s-AGCN~\citep{Shi2019TwoStreamAG} network is used as the online and target networks for the BYOL based learning.
	\item ST-GCN~\citep{yan2018spatial} + BYOL: ST-GCN~\citep{yan2018spatial} network is used as the online and target networks for the BYOL based learning.
\end{itemize}

The results of comparison are listed in Table~\ref{a}. From the results, it can be seen that ST-GCN~\citep{yan2018spatial} performs better than other supervised methods~\citep{Liu2020DisentanglingAU,Chen2021ChannelwiseTR,Shi2019TwoStreamAG}, and the proposed online network further outperforms ST-GCN~\citep{yan2018spatial} and achieves the best performance in the BYOL based learning. Moreover, ablation experiment on different structural compositions of our feature transformation network is also conducted. Different layers of ST-GCN and GConv-GRU are used to construct different versions of the online network, including
\begin{itemize}
	\item v1: 8 ST-GCN layers + 1 GConv-GRU layer
	\item v2: 6 ST-GCN layers + 1 GConv-GRU layer
	\item v3: 2 ST-GCN layers + 1 GConv-GRU layer
	\item v4: 4 ST-GCN layers + 0 GConv-GRU layer (temporal pooling instead)
	\item v5: 4 ST-GCN layers + 2 GConv-GRU layer
	\item Proposed: 4 ST-GCN layers + 1 GConv-GRU layer
\end{itemize}

The results of comparison are also listed in Table~\ref{a}. It can be seen that the proposed online network with 4 ST-GCN layers + 1 GConv-GRU layer performs the best. It can also be seen that when increasing the layers of ST-GCN or GConv-GRU over the proposed one, the performance can no longer be improved. This behaves differently to the supervised learning networks such as the ST-GCN~\citep{yan2018spatial} with deep layers, indicating that it tends to be overfitting for unsupervised skeleton action recognition learning as described in Section \ref{sec_motivation}. Moreover, it cannot extract effective features with too few layers of ST-GCN such as online network v3. This validates the effectiveness of our feature transformation network in extracting the spatial-temporal features and in reducing the overfitting (with less parameters than ST-GCN~\citep{yan2018spatial}). Also from the perspective of computation, in practical use, only the proposed online network and the final output layer for recognition is needed and thus takes less complexity than the supervised ST-GCN~\citep{yan2018spatial}. Therefore, four ST-GCN layer and one GConv-GRU layer is used for the proposed online network in the following experiments.

%{\color{red} Moreover, ablation experiment on different structural compositions of our feature transformation network is also conducted. Different layers of ST-GCN and GConv-GRU are used to build different versions of the online network, including 8 ST-GCN layers + 1 GConv-GRU layer (v1),  6 ST-GCN layers + 1 GConv-GRU layer (v2), 2 ST-GCN layers + 1 GConv-GRU layer (v3), 4 ST-GCN layers without GConv-GRU (temporal pooling instead) (v4), 4 ST-GCN layers + 1 GConv-GRU layer performs the best. It can also be seen that when increasing the layers of ST-GCN or GConv-GRU over the proposed one, the performance can no longer be improved. This behaves differently to the supervised learning networks such as the ST-GCN~\citep{yan2018spatial} with deep layers, indicating that it tends to be overfitting for unsupervised skeleton action recognition learning as described in Section \ref{sec_motivation}. Moreover, it will not extract effective features for using too few layers of ST-GCN such as online network v3. This validates the effectiveness of our feature transformation network in extracting the spatial-temporal features and in reducing the overfitting (with less parameters than ST-GCN). The proposed online network is constructed with four ST-GCN layer and one GConv-GRU layer in the following experiments.}

%  due to overfitting %Furthermore, the proposed UST-FEFP can further improve performance.

{\bf{Evaluation of combining BYOL based learning and pretext task based learning}}: As discussed in the Motivation, \textit{rich distributed spatial-temporal} features \textit{containing all information of the original skeleton} need to be generated in unsupervised learning. In order to validate this, the proposed U-FEFP is compared with the two separate modules, proposed online network with BYOL based learning and proposed online network with pretext task based learning. The results are shown in Table \ref{exp_modules}. %{\color{red} It can be seen that proposed online network outperforms ST-GCN in pretext task and proposed skeleton reconstruction network is better than skeleton reconstruction network1. }
The proposed U-FEFP combining the BYOL and pretext task based learning outperforms the two separate modules, validating they can complement each other. Moreover, the results of BYOL based learning significantly outperforms the pretext task based learning, validating our argument in Motivation that rich distributed features in unsupervised learning matters the most since skeleton is already high-level and low-dimension features.

{\bf{Evaluation of the overfitting under different methods}}: In order to illustrate the overfitting problem described in Section \ref{sec_motivation}, the unsupervised pre-training, fine-tuning and test processes of the proposed U-FEFP and 2s-AGCN are visualized as shown in Fig. \ref{loss}. The fine-tuning and test processes refer to the fine-tuning and test in the linear evaluation protocol where only one last fully connected layer is trained. In Fig. \ref{loss}, the unsupervised pre-training process is characterized in terms of loss while the fine-tuning and test processes are in terms of accuracy and the test accuracy is achieved for each epoch in the fine-tuning process. By comparing the unsupervised pre-training and fine-tuning/test in Fig. \ref{loss}, it can be seen that while 2s-AGCN achieves much smaller loss in the unsupervised pre-training stage, it performs significantly worse than the proposed U-FEFP in the fine-tuning/test in the linear evaluation protocol stage, validating its overfitting. Moreover, by comparing the fine-tuning and test in Fig. \ref{loss}, it can be seen that the accuracy gap between fine-tuning and test of 2s-AGCN is much larger that of the proposed U-FEFP. While the 2s-AGCN is fine-tuned to a relatively better performance, the gap is also becoming much larger. By contrast, the performance of the proposed U-FEFP is significantly better than 2s-AGCN with a very small gap between fine-tuning and test, and only gets slightly increased in the fine-tuning process. This demonstrates that the proposed U-FEFP is much less prone to overfitting compared to the existing methods. 

{\bf{Evaluation of different target decay rates}}: The proposed U-FEFP is also evaluated with different target decay rates used in the BYOL based learning, and results are shown in Table~\ref{bb}. It can be seen that proposed U-FEFP performs better when $\tau$ is 0.99. Therefore,  target decay rate is set to 0.99 in our experiments.

\begin{table}[htb]
	\begin{center}
		\caption{Comparison of different target decay rates } \label{bb}
		\begin{tabular}{l  c c } \hline		
			$\tau$ & X-Sub (\%) & X-View (\%) \\  \hline 
			0.9 & 80.84& 85.92\\
			0.99 & 82.50& 87.52\\
			0.999 & 80.86& 86.14\\
			\hline 		
		\end{tabular}
	\end{center}
\end{table}

\begin{table}[t]
	\begin{center}
		\caption{Comparison of different batch sizes} \label{b}
		\begin{tabular}{l  c c } \hline		
			Batch-size& X-Sub (\%) & X-View (\%) \\  \hline 
			64 & 79.72& 84.54\\
			128 & 80.31& 85.92\\
			256 & 82.10& 87.04\\
			512  & 82.50&  87.52\\
			1024 & 82.63&  87.66\\
			\hline 		
		\end{tabular}
	\end{center}
\end{table}

\begin{table}[htb]
	\setlength{\tabcolsep}{2pt}
	\begin{center}
		\caption{ Comparison of different methods in the semi-supervised setting on NTU RGB+D 60 dataset.} \label{c}
		\begin{tabular}{l  c c  c} \hline		
		\multirow{2}{*}{Method}	& \multirow{2}{*}{\makecell[c]{Label\\ fraction(\%)} } &\multirow{2}{*}{\makecell[c]{X-Sub\\ (\%)}}  & \multirow{2}{*}{\makecell[c]{X-View \\(\%) }}  \\
		& & & \\
		  \hline  
	
			LongT GAN~\citep{Zheng2018UnsupervisedRL} &1& 35.20&-\\
			MS$^{2}$L~\citep{Lin2020MS2LMS} &1& 33.10&-\\
			ISC~\citep{Thoker2021SkeletonContrastive3A}& 1& 35.70 &38.10\\
			SKT~\citep{Zhang2022SkeletalTU}&1 & 43.20&44.90\\
			SRCL~\citep{Zhang2022UnsupervisedSA} &1 & 52.60 & 53.30\\
			\textbf{U-FEFP}  &1 & \textbf{56.20}& \textbf{57.90}\\
			\hline 
			LongT GAN~\citep{Zheng2018UnsupervisedRL} &10& 62.00&-\\
			MS$^{2}$L~\citep{Lin2020MS2LMS} &10& 65.20&-\\
			ISC~\citep{Thoker2021SkeletonContrastive3A}& 10& 65.90&72.50\\
			SKT~\citep{Zhang2022SkeletalTU}&10 & 67.60&71.30\\
			SRCL~\citep{Zhang2022UnsupervisedSA} &10 & 69.30  &76.20\\	
			\textbf{U-FEFP}  &10 & \textbf{73.80}  &\textbf{79.40}\\
			\hline 	
		\end{tabular}
	\end{center}
\end{table}

{\bf{Evaluation of different batch sizes}}: The proposed U-FEFP is also evaluated with different batch sizes and results are shown in Table~\ref{b}. It can be seen that using larger batch size performs better, because the variation in a large batch size may improve the BYOL based learning performance. However, the improvement is limited using 1024 batch-size which requires large GPU memory. Therefore, batch size is set to 512 in our experiments.

{\bf{Evaluation of semi-supervised learning}}: The proposed U-FEFP is also verified in the semi-supervised learning way. Firstly, the online network is trained by unsupervised manner and  fine-tuned with 1\% and 10\% labeled data, respectively. The results compared with the existing methods are shown in Table~\ref{c}. It can be seen that compared with other unsupervised learning methods (in the same semi-supervised learning setting), the proposed U-FEFP also achieves better performance. This validates that the proposed U-FEFP can work in different training settings and perform better than others.%The effectiveness of our proposed UST-FEFP is verified in semi-supervised learning.

\begin{figure}[tb]
	\centering
	\includegraphics[width=0.5\textwidth,height=0.5\textwidth]{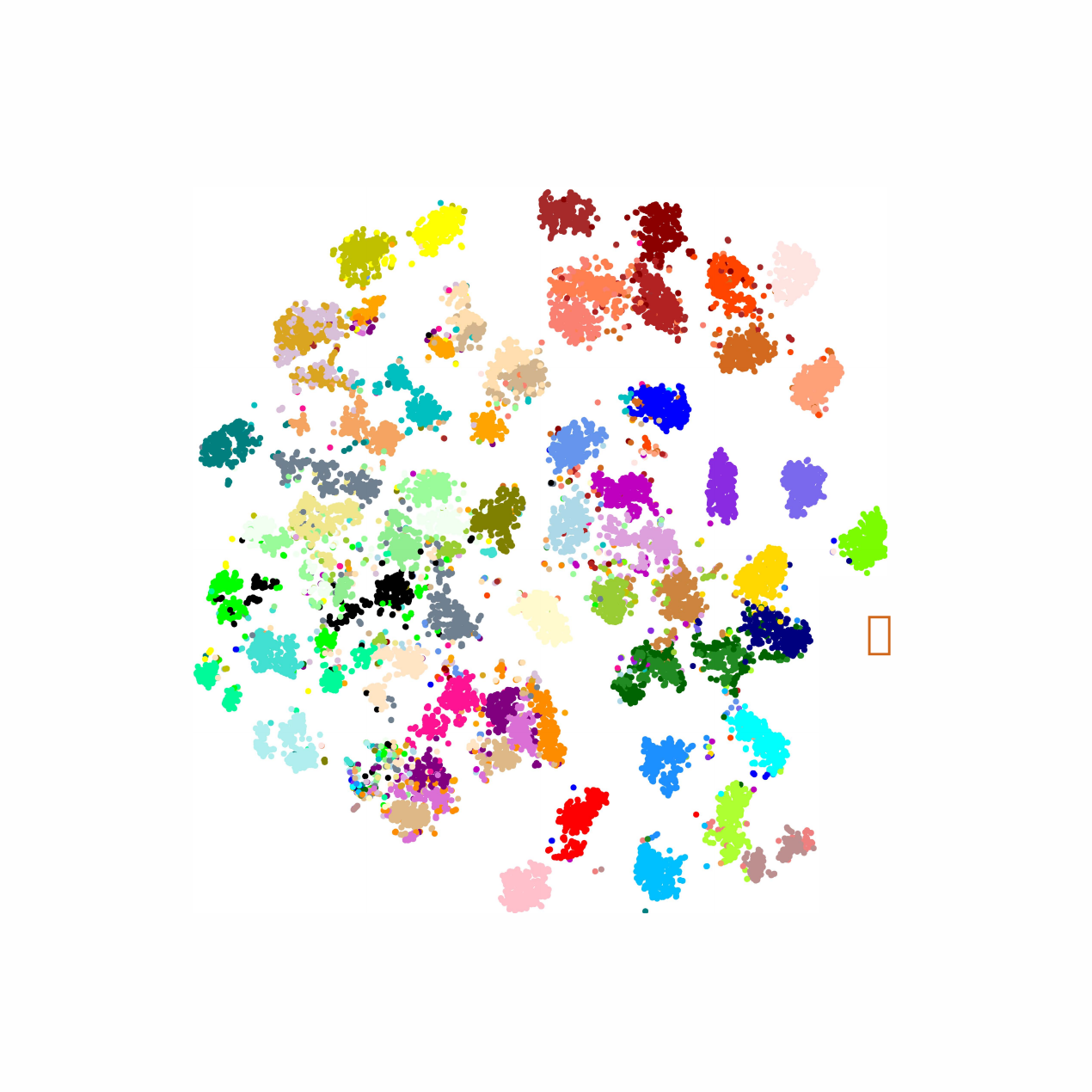}
	\caption{t-SNE visualization of embedding for U-FEFP on the NTU-60 X-View	task }\label{sne}  
\end{figure}

	\begin{table}[htb]
	\setlength{\tabcolsep}{1pt}
	
	\begin{center}
		\caption{ Experimental results (accuracy)  on the NTU-60. }  \label{d}
		\begin{tabular}{l  c c c } \hline		
			\multirow{2}{*}{Method}& \multirow{2}{*}{\makecell[c]{Train\\ manner}}  &\multirow{2}{*}{\makecell[c]{X-Sub\\ (\%)} } &\multirow{2}{*}{\makecell[c]{X-View \\(\%)}} \\ 
		
			& & &  \\  \hline 
			Lie group~\citep{Vemulapalli2014HumanAR} &supervised & 50.10 & 52.80\\
			H-RNN~\citep{Du2015HierarchicalRN} &supervised & 59.10 &64.00\\
			PA-LSTM~\citep{shahroudy2016ntu} &supervised & 62.90 &70.30\\
			ST-LSTM+TS~\citep{Liu2017SkeletonBasedAR} &supervised & 69.20 &77.70\\
			STA-LSTM~\citep{Song2017AnES} &supervised & 73.40  &81.20\\
			Visualize CNN~\citep{Liu2017EnhancedSV} &supervised & 76.00 &82.60\\
			C-CNN+MTLN~\citep{Ke2017ANR} &supervised & 79.60 &87.70\\
			VA-LSTM~\citep{Zhang2017ViewAR} &supervised & 79.20& 88.30\\
			IndRNN~\citep{Li2018IndependentlyRN} &supervised & 81.80 &88.00\\
			ST-GCN~\citep{yan2018spatial} &supervised & 81.50&88.30\\
			\hline 
			LongT GAN~\citep{Zheng2018UnsupervisedRL} &unsupervised& 39.10 &48.10\\
			PCRP~\citep{Xu2020PrototypicalCA} &unsupervised& 53.90 &63.50\\
			ASCAL~\citep{Rao2021AugmentedSB} &unsupervised& 58.50 &64.80\\
			MS$^{2}$L~\citep{Lin2020MS2LMS} &unsupervised& 52.60 &-\\
			P\&C~\citep{Su2020PREDICTC}& unsupervised& 50.70&76.30\\
			CRRL~\citep{Wang2022ContrastReconstructionRL}& unsupervised& 67.60 &73.80\\
			SKT~\citep{Zhang2022SkeletalTU}&unsupervised & 72.60 &77.10\\
			ISC~\citep{Thoker2021SkeletonContrastive3A}& unsupervised& 76.30 &85.20\\
			CrossSCLR~\citep{Li20213DHA}&unsupervised & 72.90 &79.90\\
			CrossSCLR (3S)~\citep{Li20213DHA}&unsupervised & 77.80 &83.40\\
			SRCL~\citep{Zhang2022UnsupervisedSA} &unsupervised & 77.30  &82.50\\
			SRCL (3S)~\citep{Zhang2022UnsupervisedSA} &unsupervised & 80.90 &85.60\\
			ST-CL~\citep{Gao2023EfficientSC} &unsupervised & 68.10 &69.40\\
			CrossMoCo~\citep{Zeng2023Contrastive3H} &unsupervised &78.40 &84.90\\
			HaLP~\citep{10203218} &unsupervised &79.70  &86.80\\
			3s-ActCLR~\citep{Lin2023ActionletDependentCL} &unsupervised & 84.30 &88.80\\
			\hline 
			U-FEFP  &unsupervised & 82.50 &87.52\\
			\textbf{U-FEFP (3S)}  &unsupervised & \textbf{86.92} &\textbf{91.44}\\
			\hline 
			
		\end{tabular}
	\end{center}
\end{table}

\subsection{Comparison with the State-of-the-Art Methods}
The proposed U-FEFP is compared with existing state-of-the-art methods on the different datasets. The comparison results on the NTU-60 are shown in Table~\ref{d}. It can be seen that the proposed U-FEFP outperforms the existing unsupervised methods~\citep{Su2020PREDICTC,Zheng2018UnsupervisedRL,Lin2020MS2LMS,Rao2021AugmentedSB,Li20213DHA,Thoker2021SkeletonContrastive3A,Zhang2022UnsupervisedSA,Zhang2022SkeletalTU,Gao2023EfficientSC,Zeng2023Contrastive3H}. The proposed U-FEFP only using joint is even better than method~\citep{Li20213DHA} using joint, bone and motion data (3S) together. The proposed U-FEFP even performs better than some supervised learning based methods~\citep{Vemulapalli2014HumanAR, Du2015HierarchicalRN, yan2018spatial, shahroudy2016ntu,Song2017AnES, Liu2017EnhancedSV,Li2018IndependentlyRN}.

\begin{figure*}[htb]
		\centering
		\includegraphics[width=0.95\textwidth]{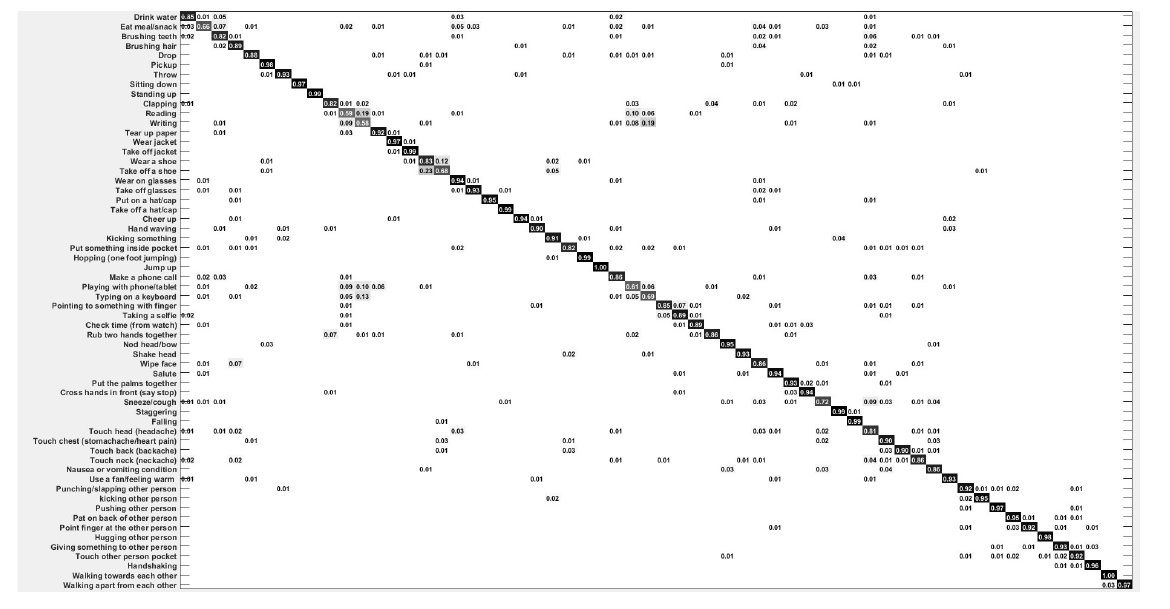}
		\caption{The confusion matrix for U-FEFP on the NTU-60 X-View.}\label{cv}  
	\end{figure*}

Furthermore, t-SNE~\citep{Maaten2008VisualizingDU} is used to visualize the embedding clustering produced by the proposed U-FEFP on the NTU-60 X-view task using all the data. The t-SNE illustration is shown in Fig.~\ref{sne}. It can be seen that proposed U-FEFP can learn more discriminative latent space. Although some samples may deviate from their action class centers, they are also away from other action classes, making them easier to be discriminated. This forms the difference between the features produced by the unsupervised learning and supervised learning, and demonstrates the importance of rich distributed features where different distributed features may be used to discriminate different samples in unsupervised learning.   Moreover, the confusion matrix for the proposed U-FEFP on the NTU-60 X-View is shown in Fig.~\ref{cv}. It can be seen that most of the action classes are recognized with high accuracy, but actions with similar small gesture motion can be confused such as ``writing'' and ``reading''.

	%,height=0.3\textwidth
	
	\begin{table}[htb]
	\setlength{\tabcolsep}{1pt}
		\begin{center}
			\caption{  Experimental results (accuracy)  on the NTU-120. }  \label{e}
			\begin{tabular}{l  c c c c c} \hline		
				\multirow{2}{*}{Method}& \multirow{2}{*}{\makecell[c]{Train\\ manner}}  &\multirow{2}{*}{\makecell[c]{X-Sub\\ (\%)}}  &\multirow{2}{*}{\makecell[c]{X-Set \\(\%)}} \\ 
		
				& &  & \\  \hline 
				PA-LSTM~\citep{shahroudy2016ntu} &supervised & 25.50  &26.30\\
				SkeMotion~\citep{Liu2017GlobalCA} &supervised & 67.70  &66.90\\
				Multi CNN~\citep{Ke2018LearningCR} &supervised & 62.20  &61.80\\
				TSRJI~\citep{Caetano2019SkeletonIR} &supervised & 67.90 &62.80\\
				ST-GCN~\citep{yan2018spatial} &supervised & 70.70 &73.20\\
				\hline
				ASCAL~\citep{Rao2021AugmentedSB} &unsupervised& 48.60 &48.60\\
				CRRL~\citep{Wang2022ContrastReconstructionRL}& unsupervised& 56.20 &57.00\\
				SKT~\citep{Zhang2022SkeletalTU}&unsupervised & 62.60 &64.30\\
				ISC~\citep{Thoker2021SkeletonContrastive3A}& unsupervised& 67.90 9.66&67.10\\
				
				CrossSCLR (3S)~\citep{Li20213DHA}&unsupervised & 67.90&66.70\\
				SRCL~\citep{Zhang2022UnsupervisedSA} &unsupervised & 67.206&67.90\\
				SRCL (3S)~\citep{Zhang2022UnsupervisedSA} &unsupervised & 71.80&72.90\\
				
				ST-CL~\citep{Gao2023EfficientSC} &unsupervised & 54.20 &55.60 \\
				HaLP~\citep{10203218} &unsupervised & 71.10  &72.20\\
				3s-ActCLR~\citep{Lin2023ActionletDependentCL} &unsupervised & 74.30 &75.70\\
				\hline 
				U-FEFP  &unsupervised & 73.85 &74.80\\
				\textbf{U-FEFP (3S)}  &unsupervised & \textbf{77.56} &\textbf{79.66}\\
				\hline 
				
			\end{tabular}
		\end{center}
	\end{table}
	
	The result comparison of the proposed U-FEFP against the existing methods on the NTU-120 dataset is shown in Table~\ref{e}. The proposed U-FEFP (3S) obtains 77.56\% and 79.66\% on X-Sub and X-Set, respectively, and achieves the state-of-the-art performance. U-FEFP using joint, bone and motion data is better than ST-GCN using supervised way, which demonstrates the effectiveness of U-FEFP.

	The result comparison on the PKU-MMD dataset is shown in Table~\ref{pku}. From the table, it can be seen that U-FEFP (3S) achieves 92.30\% and 57.80\% on PKU-MMD I and PKU-MMD II, respectively. Compared with the previous best method (i.e., SRCL \citep{Zhang2022UnsupervisedSA}), the proposed method improves by 4.10\% and 4.60\% on PKU-MMD I and PKU-MMD II, respectively. The performance is significantly improved compared with the supervised ST-GCN~\citep{yan2018spatial}, demonstrating the effectiveness of the proposed U-FEFP.
	
	\begin{table}[htb]
		\setlength{\tabcolsep}{2pt}
		\begin{center}
			\caption{ Experimental results (accuracy) on the PKU-MMD. } \label{pku}
			\begin{tabular}{l  c c c c c  } \hline	
				\multirow{2}{*}{Method}& \multirow{2}{*}{\makecell[c]{Train\\ manner}}  &\multirow{2}{*}{\makecell[c]{part I \\ (\%)}}  &\multirow{2}{*}{\makecell[c]{part II \\(\%)} }\\ 
				
				& &  & \\  \hline 	
				
				ST-GCN~\citep{yan2018spatial} &supervised & 84.10 &  48.20\\
				\hline 
				LongT GAN~\citep{Zheng2018UnsupervisedRL} &unsupervised& 67.70& 27.00\\
				
				MS$^{2}$L~\citep{Lin2020MS2LMS} &unsupervised& 64.90& 27.60\\

				ISC~\citep{Thoker2021SkeletonContrastive3A}& unsupervised& 80.90& 36.00\\
				CRRL~\citep{Wang2022ContrastReconstructionRL}& unsupervised& 82.10& 41.80\\
				CrossSCLR (3S)~\citep{Li20213DHA}&unsupervised & 84.90& -\\
				SRCL~\citep{Zhang2022UnsupervisedSA} &unsupervised & 87.40&  48.10\\
				SRCL (3S)~\citep{Zhang2022UnsupervisedSA} &unsupervised & 88.20&  53.20\\
				3s-ActCLR~\citep{Lin2023ActionletDependentCL} &unsupervised & 90.00&  55.90\\
				HaLP~\citep{10203218} &unsupervised & -&  43.50\\
				\hline 
				U-FEFP &unsupervised & 90.72& 53.12\\
				\textbf{U-FEFP (3S)}  &unsupervised & \textbf{92.30} & \textbf{57.80} \\
				\hline 
				
			\end{tabular}
		\end{center}
	\end{table}
	
	\section{Conclusion}	
	\label{sec_conclusion}
	In this paper, we propose the U-FEFP learning framework for unsupervised skeleton based action recognition. The U-FEFP produces rich distributed features containing all information of the skeleton sequence, which is vital for the unsupervised skeleton based action recognition. A relatively small spatial-temporal feature transformation subnetwork combining ST-GCN and GConv-GRU is proposed to effectively capture the skeleton sequence features. Based on this subnetwork, the unsupervised BYOL based feature enrichment learning and unsupervised pretext task based fidelity preservation learning is combined to formulate our U-FEFP, in order to produce the desired features. t-SNE is used to illustrate the features of the proposed U-FEFP, which demonstrates its advantage over the existing methods. Extensive experiments are conducted on NTU-60, NTU-120 and PKU-MMD dataset, where the proposed U-FEFP  outperforms the current state-of-the-art methods and achieves the best performance. Ablation study on the proposed modules is also performed and validates their effectiveness.

\section{Acknowledgements}
This work was partly supported by the National Natural Science Foundation of China (No. 62101512, 62001429,62271453 and 62271290), Fundamental Research Program of Shanxi Province (20210302124031) and Shanxi Scholarship Council of China (2023-131).

\bibliographystyle{model2-names}
\bibliography{st2022}

%% The Appendices part is started with the command \appendix;
%% appendix sections are then done as normal sections
%% \appendix

%% \section{}
%% \label{}

%% If you have bibdatabase file and want bibtex to generate the
%% bibitems, please use
%%
%%  \bibliographystyle{elsarticle-num}
%%  \bibliography{<your bibdatabase>}

%% else use the following coding to input the bibitems directly in the
%% TeX file.

%\begin{thebibliography}{00}

%% \bibitem{label}
%% Text of bibliographic item

%\bibitem{}

%\end{thebibliography}
\end{sloppypar}
\end{document}